\pdfoutput=1

\documentclass[11pt]{article}
\usepackage{fancyvrb}
\usepackage{listings} 
\usepackage[preprint]{stylefiles/acl}
\usepackage{times}
\usepackage{latexsym}
\usepackage[T1]{fontenc}
\usepackage[utf8]{inputenc}
\usepackage{microtype}
\usepackage{inconsolata}
\usepackage{graphicx}
\usepackage{graphicx} 
\usepackage{adjustbox}
\usepackage{amsmath}
\usepackage{amssymb}
\usepackage{bbding}
\usepackage{booktabs}
\usepackage{caption}
\usepackage{cleveref}
\usepackage{comment}
\usepackage{enumitem}
\usepackage{float}
\usepackage{graphicx}
\usepackage{hyperref}
\usepackage{makecell}   
\usepackage{listings}
\usepackage{mdframed}
\usepackage{multirow}
\usepackage{placeins}
\usepackage{scalerel}
\usepackage{soul}
\usepackage{verbatim}
\usepackage{spverbatim}
\usepackage{stfloats}
\usepackage{subcaption}
\usepackage[inkscapeformat=png]{svg}
\usepackage{tabularx}
\usepackage{tabularray}
\usepackage{tcolorbox}
\usepackage{tikz}
\usepackage{times}
\usepackage[textsize=tiny]{todonotes}
\usepackage[normalem]{ulem}
\usepackage{url}
\usepackage{xcolor}
\usepackage{xspace}
\usepackage{listings}
\usepackage{fancyvrb}

\sethlcolor{yellow} 
\colorlet{lightred}{red!45} 
\colorlet{lightgreen}{green!30}
\newcommand{\hlred}[1]{{\sethlcolor{lightred}\hl{#1}}}
\newcommand{\hlblue}[1]{{\sethlcolor{cyan}\hl{#1}}}
\newcommand{\hlgreen}[1]{{\sethlcolor{lightgreen}\hl{#1}}}

\title{Unsupervised Human Preference Learning}
\author{
Sumuk Shashidhar \quad Abhinav Chinta \quad Vaibhav Sahai \quad Dilek Hakkani-T\"ur \\ \normalfont{School of Computer Science} \\ \normalfont{University of Illinois Urbana-Champaign} \\ \texttt{\{sumuks2, achinta3, sahai3, dilek\}@illinois.edu}
}

\begin{document}
\maketitle
\begin{abstract}
Large language models demonstrate impressive reasoning abilities but struggle to provide personalized content due to their lack of individual user preference information. Existing methods, such as in-context learning and parameter-efficient fine-tuning, fall short in capturing the complexity of human preferences, especially given the small, personal datasets individuals possess. In this paper, we propose a novel approach utilizing small parameter models as preference agents to generate natural language rules that guide a larger, pre-trained model, enabling efficient personalization. Our method involves a small, local "steering wheel" model that directs the outputs of a much larger foundation model, producing content tailored to an individual's preferences while leveraging the extensive knowledge and capabilities of the large model. Importantly, this personalization is achieved without the need to fine-tune the large model. Experimental results demonstrate that our technique significantly outperforms baseline personalization methods. By allowing foundation models to adapt to individual preferences in a data and compute-efficient manner, our approach paves the way for highly personalized language model applications.
\end{abstract}
\section{Introduction}

Large Language Models have revolutionized various domains with their impressive capabilities in reasoning, knowledge retrieval, and content generation. However, a crucial aspect where they often fall short is the ability to adapt their outputs to individual user preferences. While LLMs excel at generating content that caters to a broad audience, they struggle to produce outputs that resonate with the unique stylistic and contextual needs of individual users \cite{woźniak2024personalized}. This limitation stems from the fact that LLMs are typically trained on massive, generic datasets, promoting neutrality and hindering their capacity to learn and adapt to individual nuances \cite{doi:10.14318/hau6.1.002}.

Existing personalization techniques, such as in-context learning (ICL) \cite{NEURIPS2020_1457c0d6} and parameter-efficient fine-tuning (PEFT) \cite{hu2021lora, dettmers2023qlora}, have shown promise in adapting LLMs to specific tasks. However, these methods encounter significant challenges when applied to the domain of human preference learning. ICL, while effective in providing general guidance, struggles to capture the intricate and often contradictory nature of human preferences, especially when relying on limited in-context examples \cite{peng2023does}. Similarly, PEFT methods, while efficient in terms of compute and storage, face difficulties in generalizing to new preferences, particularly when users only possess small, personal datasets \cite{balne2024parameter}. This raises a fundamental question: \textbf{how can we efficiently and effectively align powerful LLMs to individual user preferences, especially when personalized data is scarce?}

To address this challenge, we propose a novel approach based on the concept of preference agents. These agents are small, locally trainable language models designed to learn and encode individual user preferences into concise natural language rules. These agents act like a small "steering wheel," guiding the output of a much larger, generic LLM towards the desired personalized style and content. This modular architecture decouples the task of preference learning from the generic LLM, allowing users to efficiently fine-tune a small agent on their personal data without the need for modifying the weights of the larger model.

Our approach represents a significant departure from conventional preference learning methods, offering a potential solution for unsupervised human preference learning. We evaluate our method across three diverse datasets encompassing human-generated content: emails, news articles and product reviews. Our results demonstrate that preference-guided LLMs significantly outperform both fine-tuning baselines and standard prompting techniques, based on automatic metrics, GPT-4o evaluation, and human judgments.

Specifically, our contributions include:

\begin{itemize}
\item A novel fine-tuning objective that leverages distilled preference information instead of traditional input-output pairs, promoting efficient learning of user preferences.
\item Empirical evidence that the use of preference agents leads to significant performance improvements – up to 80\% in some cases – over existing personalization methods, particularly when aligning LLMs to individual styles and preferences.
\item The release of three large, human intent annotated preference datasets to foster future research in personalization. \footnote{\href{https://huggingface.co/datasets/preference-agents/}{https://huggingface.co/preference-agents}}
\end{itemize}
\section{Method}

\begin{figure*}[t!]
    \centering
    \includegraphics[width=\textwidth]{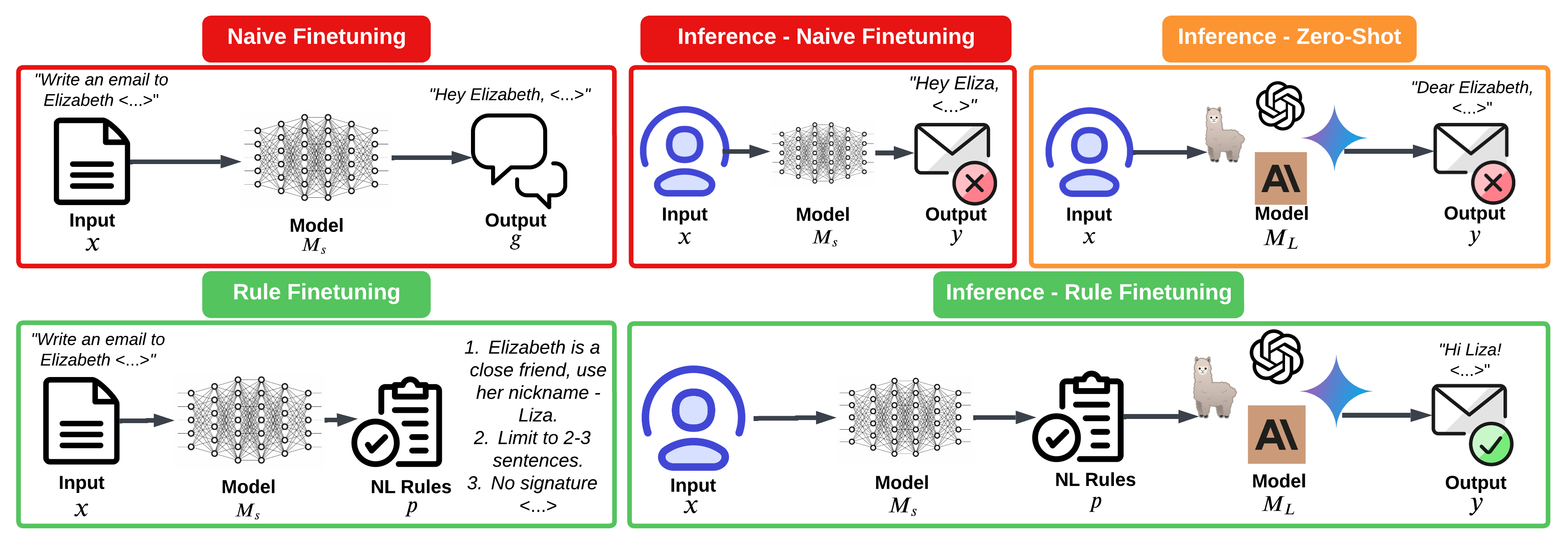}
     \caption{Preference Rule Finetuning vs Naive Finetuning and Large Model Zero-Shot}
    \label{fig:pref-agents-teaser}
\end{figure*}

In this section, we detail our approach for aligning language models to personalized user preferences using small preference agents. Our method involves two key components: generating natural language rules that capture user preferences and utilizing these rules to guide a larger, pre-trained language model. This modular architecture allows for efficient personalization without extensive retraining.

\subsection{Task Definition}

Given a task \( T \), we define the dataset \( \mathcal{D} \) as consisting of input-output pairs. Each input comprises a user intent \( \mathbf{u} \) and associated task metadata \( \mathbf{m} \), and the output is the ideal task completion, denoted as \( \mathbf{g} \), which we consider the ground truth. Thus, the dataset can be formally expressed as:

\[
\mathcal{D} = \{(\mathbf{x}, \mathbf{g}) \mid \mathbf{x} = (\mathbf{u}, \mathbf{m})\}
\]

\subsection{Constraints and Assumptions}

We seek to enable users to generate high quality, personalized responses as our goal, which are bounded by some constraints and assumptions:
\begin{itemize}
    \item \textbf{Constraint 1:} The size of the dataset $\mathcal{D}$ is not large enough to permit effective full parameter model fine-tuning. Given that individual users typically possess small, personal datasets, it is impractical to expect these datasets to be sufficient for extensive fine-tuning of a large language model.
    \item \textbf{Constraint 2:} The small model, denoted as $M_S$, must be lightweight enough to operate (w.r.t both training and inference) on lower power end-user devices. This requirement ensures that users can generate and apply their preferences without the need for high-performance computing resources. This allows for local inference, making the personalization process more accessible and convenient.
    \item \textbf{Constraint 3:} The large model, referred to as $M_L$, is either too large to run inference locally or is a closed-source API model. Consequently, it is not feasible, or cost effective to fine-tune or align $M_L$ by altering its model weights.
\end{itemize}

\subsection{Model Training}
\label{section:method/model-training}

Given the dataset $\mathcal{D}$, we first task $M_L$ with generating zero-shot responses to our training data. These initial responses are devoid of any user-specific preference information:

\begin{equation}
\mathbf{Y}_z = M_L(\mathbf{X})
\end{equation}

\noindent
where $\mathbf{Y}_z$ represents the set of zero-shot outputs for all inputs $\mathbf{X}$ in the training dataset.

Next, we leverage $M_L$'s capabilities to extract the delta between the zero-shot completions ($\mathbf{Y}_z$) and the ground truth outputs ($\mathbf{G}$). This delta represents the preference rules that need to be learned by the smaller model:

\begin{equation}
\mathbf{P} = M_L(\mathbf{Y}_z, \mathbf{G})
\end{equation}

Here, $\mathbf{P}$ represents the set of preference rules derived for each training example.  We hypothesize that $M_L$ can effectively identify these rules without prior knowledge of the specific user's preferences, just by observing the differences between the zero shot completion and the ground truth.

Finally, we train the smaller model, $M_S$, to learn to generate these preference rules.  The training data for $M_S$ consists of input-preference rule pairs:

\begin{equation}
M_S \xrightarrow{(\mathbf{X}, \mathbf{P})} M_A
\end{equation}

Through this training process, $M_S$ learns to map user intents and task metadata to natural language preference rules, effectively becoming a personalized preference agent ($M_A$).

\subsection{Model Alignment}

Once the preference agent $M_A$ is trained, we can use it to align the larger model's outputs to unseen user data. For a new input $\mathbf{x}$, we first generate preference rules using the trained agent:

\begin{equation}
\mathbf{p} = M_A(\mathbf{x})
\end{equation}

These rules, expressed in natural language, are then provided as additional context to the large language model $M_L$ alongside the original input:

\begin{equation}
y_a = M_L(\mathbf{x}, \mathbf{p})
\end{equation}

The output $y_a$ is considered to be preference-aligned as it is generated by $M_L$ while considering the user's preferences encoded in $\mathbf{p}$. This approach allows us to leverage the vast knowledge of $M_L$ while tailoring the output to individual preferences without directly modifying the large model's weights.

\subsection{Quantifying Alignment}

We utilize an evaluation function $\text{Eval}(y_a, y_z | \mathbf{x})$ on an unseen test set $\mathcal{T}$. For each example in $\mathcal{T}$, the function compares the preference-aligned output $y_a$ with the zero-shot output $y_z$ generated by $M_L$ without preference rules. The evaluation function assigns a score indicating the preference between $y_a$ and $y_z$, given the input $\mathbf{x}$. A positive score indicates a preference for the aligned output $y_a$, suggesting better alignment with the user's likely preference, while a negative score favors the zero-shot output. We aggregate these scores across all examples in $\mathcal{T}$ to obtain an overall alignment score:

\begin{equation}
\text{Score}(\mathcal{T}) = \sum_{i=1}^{|\mathcal{T}|} \text{Eval}(y_{a}^{(i)}, y_{z}^{(i)} | \mathbf{x}^{(i)})
\end{equation}

where $|\mathcal{T}|$ represents the number of examples in the test set and $y_{a}^{(i)}$ and $y_{z}^{(i)}$ represent the aligned and zero-shot outputs, respectively, for the $i$-th example.

A positive $\text{Score}(\mathcal{T})$ indicates that the preference agent successfully guides the LLM to generate outputs better aligned with user preferences compared to the baseline zero-shot outputs.
\section{Experimental Setup}

\subsection{Model Choice}

We employ Llama-3-8B-Instruct as our smaller, locally inferrable model ($M_S$) due to its strong capabilities and suitability for QLoRA fine-tuning on consumer hardware \cite{dettmers2023qlora}. For our larger reasoning model ($M_L$), we utilize Llama-3-70B-Instruct \cite{dubey2024llama3herdmodels}, Claude 3.5 Sonnet \cite{anthropic2024claude35}, and Gemini 1.5 Pro \cite{geminiteam2024gemini15unlockingmultimodal}. This diverse selection allows us to evaluate our approach with both a strong open-source model (Llama-3-70B-Instruct) and powerful closed-source API models (Claude 3.5, Gemini 1.5), to demonstrate generalizability.

\subsection{Datasets}

Our evaluation spans three datasets encompassing single and multi-user preference information:

\textbf{Enron Email Corpus.} For evaluating short-form writing personalization, we utilize the Enron email corpus \cite{10.1007/978-3-540-30115-8_22}, comprising emails from approximately 150 users, primarily senior management at Enron. We sample 15 users to analyze the reproducibility of individual writing styles. Each user's subset is split into an 80-20 train-test split.

\textbf{New Yorker.} To assess performance on long-form creative writing, we employ a subset of the All the News 2.0 dataset \cite{allthenews2}, specifically articles from The New Yorker magazine. This subset, containing approximately 3,500 articles, provides a rich source of author preference information. We investigate whether our preference agents can reproduce the unique style of The New Yorker using natural language rules. We split this dataset into a 50-50 train-test split.

\textbf{Amazon Review Subset (LAMP 3U)}. To further assess the generalizability of our framework beyond long-form content, we incorporate the LAMP 3U dataset \cite{salemi2024lamplargelanguagemodels}, which consists of Amazon product reviews grouped by user. We sampled 15 random users from LAMP 3U and then generated review intents and rules for each user following the same process employed for the Enron dataset.

Refer to Appendix \ref{appendix:dataset-enron} for details regarding dataset preparation and sampling.

\begin{table}[ht!]
\centering
\scalebox{0.8}{
\begin{tabular}{@{}l>{\raggedleft\arraybackslash}p{2.5cm}@{}}
\toprule
\textbf{Metric} & \textbf{Value} \\
\midrule
\multicolumn{2}{@{}c}{\textit{\textbf{Enron-42K (Short Form)}}} \\
\cmidrule(l){1-2}
Number of Data Points & 40,240 \\
Number of Unique Senders & 191 \\
Avg. Token Count (Email Content) & 58.83 \\
Avg. Token Count (Previous Context) & 261.48 \\
\midrule
\multicolumn{2}{@{}c}{\textit{\textbf{LAMP 3U (Medium Form)}}} \\
\cmidrule(l){1-2}
Number of Data Points & 22,500 \\
Number of Unique Users & 15 \\
Avg. Token Count (Review) & 144.32 \\
\midrule
\multicolumn{2}{@{}c}{\textit{\textbf{New Yorker (Long Form)}}} \\
\cmidrule(l){1-2}
Number of Data Points & 1,525 \\
Number of Unique Article Authors & 401 \\
Avg. Token Count (Article) & 846.60 \\
\bottomrule
\end{tabular}
}
\caption{Dataset Statistics - Enron, New Yorker, and LAMP 3U Amazon Reviews}
\label{tab:dataset_comparison}
\end{table}

\subsection{Dataset Augmentation}

\subsubsection{Synthetic Intent Generation}
We aimed to develop a fully unsupervised approach that avoids manual collection of human intents, which can be time-consuming and costly. To achieve this, we leveraged the large language model ($M_L$) to automatically extract the core content of each email or article into bullet points, emulating user input or intent.
To ensure the quality of these synthetic intents, we randomly sampled a subset and subjected them to manual human evaluation. Our findings indicated a high degree of fidelity, with over 95\% of the synthetic intents rated highly by humans.
To introduce variance and simulate different user styles, we generated three intent variants for each data point at temperatures of \texttt{0.7}, \texttt{1.0}, and \texttt{1.2}. These intents were then randomly sampled to create intent-annotated versions of our datasets. Examples of generated intents can be found in Appendix \ref{appendix:generated_intents}.

\subsubsection{Rule Generation}

To capture the nuanced stylistic preferences inherent in our datasets, we employed the large reasoning model ($M_L$) to generate natural language preference rules. These rules distill the essence of the desired style by highlighting the discrepancies between a zero-shot baseline generated by $M_L$ and the ground truth email or article.

We investigated three distinct strategies for generating these preference rules:

\begin{itemize}
    \item \textbf{Distillation-Based Rule Generation}: This approach leverages a distillation process. $M_L$ first generates a zero-shot baseline response for the given input. By analyzing the differences between this baseline and the ground truth, the model identifies missing stylistic and preference elements and generates targeted rules to bridge the gap. 
    
    \item \textbf{Direct Rule Generation}: We also explore directly prompting $M_L$ to generate rules based solely on the ground truth email or article. This approach, while simpler, lacks the targeted feedback mechanism inherent in the distillation-based method.
    
    \item \textbf{Rule Generation with Thinking Tokens}: To further enhance the rule generation process, we incorporate "thinking tokens" into the prompts. These tokens encourage the model to engage in more deliberate reasoning before generating rules, potentially leading to more insightful and effective guidance.
\end{itemize}

Examples of generated rules for each strategy are provided in Appendix~\ref{appendix:generations/rules}. A detailed discussion of the relative merits and limitations of these strategies is presented in Appendix~\ref{appendix:rulegen-strategies}.

\subsection{Preference Agent Training}

The preference agents, based on Llama-3-8B-Instruct, are trained using Quantized Low-Rank Adaptation (QLoRA) \cite{dettmers2023qlora}, a parameter-efficient fine-tuning (PeFT) method. We choose QLoRA over full fine-tuning due to its scalability and feasibility for local deployment on user devices. All model training procedures are designed to be accommodated within 16GB of VRAM, making our approach accessible to standard consumer-grade devices.

We employ a consistent set of hyperparameters across all experiments. This simplified configuration, while not optimal, serves to demonstrate the effectiveness of our method even with straightforward hyperparameter choices.  A detailed analysis of our fine-tuning procedure, including further exploration of hyperparameter search and impact on performance, can be found in Appendix \ref{appendix:finetuning-analysis}.

To establish a performance baseline, we also nai-ve finetune a model ($M_F$) using the same setup. This model is trained directly on input-output pairs (user intent and task metadata as input, ground truth text as output). This ensures a fair comparison by isolating the impact of our proposed rule-based fine-tuning approach from potential differences in model architecture or training configurations.

\subsection{Evaluation Methodology}
\label{experimental:eval-metrics}

To rigorously assess the efficacy of our preference alignment approach, we employ two complementary evaluation methodologies: automated evaluation using GPT-4 Omni (GPT-4o) and human evaluation.

\textbf{Baselines.} We compare our proposed preference agent models against several strong baselines, such as
\begin{itemize}
    \item \textbf{Zero-shot generations} from both the small model (Llama-3-8B, $M_S$) and the large model (Llama-3-70B, $M_L$),
    \item \textbf{Few-shot generations} using $M_L$, providing a limited number of examples in the prompt,
    \item \textbf{Naive fine-tuned agent} ($M_F$), where $M_S$ is directly fine-tuned on input-output pairs using QLoRA with the same hyperparameters as our preference agent training.
\end{itemize}

\textbf{Automated Evaluation.} We leverage GPT-4o as our automated evaluator due to its demonstrated capabilities in assessing human-written text and capturing stylistic nuances \cite{naismith-etal-2023-automated, zheng2023judging, shashidhar-etal-2023-democratizing}. Our primary automated metric is the win percentage, which quantifies how often a method's output is selected by GPT-4o as the best match to the ground truth. GPT-4o's evaluations are based on criteria such as similarity in style, tone, characteristic phrases, and overall resemblance to the ground truth content.

\textbf{Human Evaluation:} We complement the automated evaluation with a human evaluation study on a subset of each dataset. Participants are presented with the original input, the ground truth output, and the outputs generated by each method. They are then asked to select the response that they believe best aligns with the ground truth, using the same criteria as the GPT-4o evaluation. This human evaluation provides valuable insights into the alignment of model outputs with human preferences. Detailed information on the human evaluation protocol can be found in Appendix \ref{appendix:human-eval}.

\textbf{Choice of Metrics.} We have deliberately chosen not to rely on traditional similarity metrics such as BLEU \cite{papineni-etal-2002-bleu} and ROUGE \cite{lin-2004-rouge}. While these metrics are valuable for assessing lexical overlap, they are less effective in capturing the nuanced aspects of stylistic similarity that are central to our evaluation. For example, consider two emails from the Enron dataset with similar content but distinct styles. One email might be written in a formal, professional tone, while the other adopts a more casual, conversational style. BLEU and ROUGE, focusing primarily on the presence or absence of specific words and phrases, might assign similar scores to both emails despite the clear difference in style perceived by a human reader. This discrepancy arises because these metrics do not adequately account for the semantic meaning and contextual usage of words, which are crucial for evaluating stylistic resemblance. This decision is further elaborated upon in Appendix \ref{appendix:automated-metrics}.

\section{Results}

We evaluated the performance of our fine-tuned preference agents against several baselines using GPT-4o and human evaluation. We report the win rates – the percentage of instances where our method outperforms the baseline – in Table \ref{table:results}.  Our baselines include zero-shot generations from both $M_S$ and $M_L$, few-shot generations using $M_L$, and a naive fine-tuned agent ($M_F$). We compare these baselines against our preference agent, trained with zero-shot baseline rules, and a no-baseline agent trained without using zero-shot information.

\begin{table*}[ht!]
    \centering
    \resizebox{\textwidth}{!}{%
        \begin{tabular}{@{}lcccccccccccc@{}}
            \toprule
            \multicolumn{1}{c}{\makecell{\textbf{Preference} \\ \textbf{Agents}}}
            & \multicolumn{3}{c}{\makecell{\textbf{New Yorker}}} 
            & \multicolumn{3}{c}{\makecell{\textbf{Enron}}} 
            & \multicolumn{3}{c}{\makecell{\textbf{LAMP 3U}}} 
            & \multicolumn{2}{c}{\makecell{\textbf{Aggregated}}} \\ 
            \cmidrule(lr){1-1}
            \cmidrule(lr){2-4} \cmidrule(lr){5-7} \cmidrule(lr){8-10} \cmidrule(lr){11-12}
            \makecell{\textbf{$M_L \to$}\\ vs $\downarrow$}
            & \makecell{\textbf{Llama3 70B} \\ \textbf{Instruct}} 
            & \makecell{\textbf{Claude 3.5} \\ \textbf{Sonnet}} 
            & \makecell{\textbf{Gemini} \\ \textbf{1.5 Pro}} 
            & \makecell{\textbf{Llama3 70B} \\ \textbf{Instruct}} 
            & \makecell{\textbf{Claude 3.5} \\ \textbf{Sonnet}} 
            & \makecell{\textbf{Gemini} \\ \textbf{1.5 Pro}} 
            & \makecell{\textbf{Llama3 70B} \\ \textbf{Instruct}} 
            & \makecell{\textbf{Claude 3.5} \\ \textbf{Sonnet}} 
            & \makecell{\textbf{Gemini} \\ \textbf{1.5 Pro}} 
            & \makecell{\textbf{LLM} \\ \textbf{Evaluation}} 
            & \makecell{\textbf{Human} \\ \textbf{Evaluation}} \\ 
            \midrule
            \makecell{Small \\ Baseline}  
            & 77.4 & 91.5 & 80.0  
            & 88.4 & 96.1 & 89.8 
            & 74.6 & 84.0 & 75.3 
            & 84.1 & 91.0 \\ 
            \midrule
            \makecell{Large \\ Baseline}  
            & 67.7 & 75.2 & 66.9 
            & 85.6 & 83.7 & 88.2 
            & 66.5 & 69.5 & 63.8 
            & 74.1 & 84.5 \\ 
            \midrule
            \makecell{Few \\ Shot}        
            & 68.3 & 62.0 & 66.7 
            & 61.1 & 68.0 & 57.4 
            & 58.3 & 57.4 & 59.4 
            & 62.0 & 73.4 \\ 
            \midrule
            \makecell{Naive \\ Finetune}  
            & 80.3 & 82.4 & 81.8 
            & 75.3 & 87.8 & 81.3 
            & 85.0 & 92.7 & 89.0 
            & 83.9 & 92.2 \\ 
            \midrule
            \makecell{No Baseline \\ Agent} 
            & 65.1 & 68.8 & 63.8 
            & 58.4 & 61.3 & 62.5 
            & 63.8 & 67.2 & 60.4 
            & 63.4 & 52.0 \\ 
            \bottomrule
        \end{tabular}
    }
    \caption{\textbf{Win Rates} of Llama3 8B $M_s$ combined with various $M_L$, evaluated by GPT4o and human evaluation.}
    \label{table:results}
\end{table*}

\textbf{Small Model Baseline.} Both GPT-4o and human evaluations agree that the small model baseline (\(M_S\)) performs significantly worse than our preference agent. This highlights the limitations of using small language models alone for tasks requiring a deep understanding of user preferences. Human evaluations show an even larger performance gap compared to GPT-4o, as humans are more adept at detecting subtle differences in style and content.

\textbf{Large Model Baseline.} While the baseline produced by the large model improves, the improvement is most noticeable in domains with lower degrees of available personalization (such as articles and product reviews), and less noticeable when direct preferences are involved (such as email writing)

\textbf{Few Shot.} While our method consistently outperforms the few-shot baseline across all datasets, the performance gap is more pronounced in the New Yorker dataset, compared to the LAMP3U or Enron datasets. We hypothesize that this difference stems from the nature of the tasks. Few-shot examples are likely more effective for email writing or product review writing, a relatively structured and concise format, than for long-form article writing, where capturing stylistic nuances requires more than a few examples. 

\textbf{Naive Finetune.} Human evaluators exhibited a stronger preference for preference agent outputs over naive fine-tuning compared to GPT-4o. Post-annotation interviews revealed that naive fine-tuning often resulted in hallucinations of crucial information, a phenomenon not effectively penalized by automated metrics but disliked by humans.

\textbf{No Baseline Agent.} The preference agent trained without access to zero-shot baseline information exhibits competitive performance, particularly when considering the marginal reduction in inference cost it offers. This makes it a viable alternative in scenarios where minimizing inference cost is a priority, even if it comes with a slight compromise in performance compared to the distillation-based approach.

Both automated and human evaluations confirm that our preference agent significantly improves the alignment of LLM outputs with individual user styles and preferences, as discussed in Section \ref{discussion:why-ft-more-effective}. The strong and consistent performance across diverse datasets and LLMs highlights the generalizability utilizing preference agents. Qualitative examples and human annotation samples of the results are provided in Appendix \ref{appendix:qualitative_examples}.
\section{Discussion}

\subsection{Enhanced Fine-Tuning through Preference Rules}
\label{discussion:why-ft-more-effective}

\begin{figure}
    \centering
    \includegraphics[width=1\linewidth]{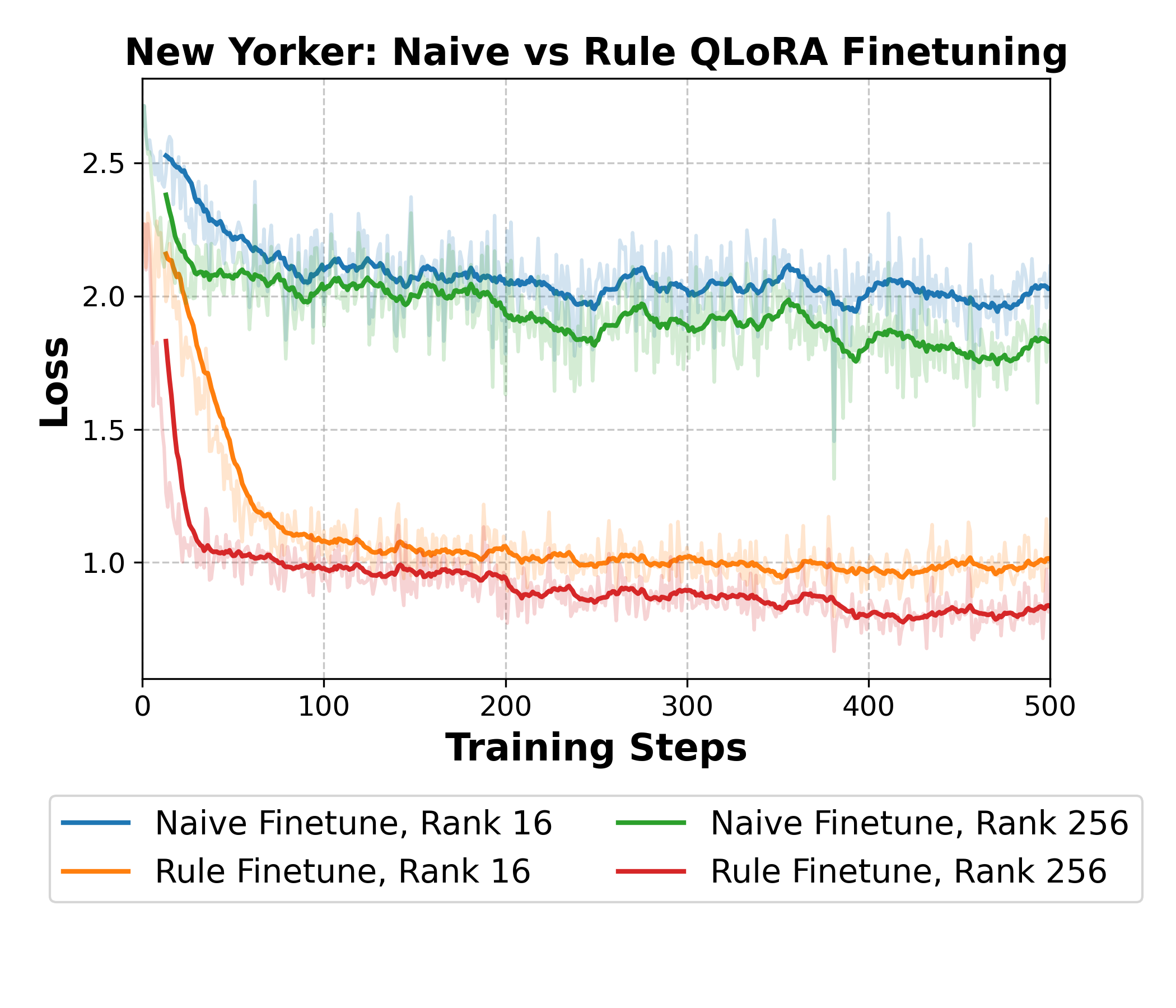}
        \caption{On the New Yorker dataset, naive fine-tuning plateaus at a loss above 1.5, whereas fine-tuning with structured preference rules reduces the loss below 1.0 with identical hyperparameters.}
    \label{fig:ruleft_vs_naiveft}
\end{figure}

Our experiments (Figure \ref{fig:ruleft_vs_naiveft}) reveal that fine-tuning the preference agent on natural language rules, as opposed to directly on input-output pairs, leads to a more effective learning process. This result indicates that structured rules provide a more efficient learning signal for the preference agent.

We hypothesize that this difference stems from the inherent complexity and diversity of the target data versus the clear structure of the preference rules. When learning from raw input-output pairs, the model must adapt to the nuances of the target task, which can be challenging given the diversity of writing styles and content. Specifically, instruction-finetuned language models often exhibit a "chatbot" style, characterized by conversational and explanatory responses \cite{ouyang2022training}. This style can be significantly different from the desired output style for specific tasks like email writing, which often requires a more direct and concise approach. Adapting the model to such specific styles directly through fine-tuning can be challenging, especially under the constraints of PeFT methods. In contrast, the structured format of the rules enables the model to discern patterns more easily, facilitating faster and more effective learning. This translates to improved sample efficiency, as the model requires fewer examples to grasp the underlying preference information.

Furthermore, this approach promotes a smaller distribution shift during fine-tuning. Naive fine-tuning necessitates a substantial adaptation to the new task's distribution, which can involve shifting the model's overall output style from a conversational "chatbot" approach to a more task-specific style. PEFT methods, while effective in adapting models to new tasks or domains, may be less effective in inducing such significant changes in the model's fundamental language generation style \cite{balne2024parameter}. On the other hand, rule-based fine-tuning focuses on learning a more specific mapping – from input to preference rules – leveraging the pre-trained language model's existing capabilities for task completion. Crucially, natural language rules, are closer to the LM's existing output distribution compared to the diverse and potentially drastically different output styles of diverse, specific tasks. This makes them more suitable for PEFT adaptation, as the model can learn to generate rules without having to undergo a substantial shift in its underlying parameters.

 his decoupling of preference learning from the core task allows for more efficient adaptation, especially in multi-task settings where the model might need to switch between different domains and writing styles. By focusing on learning user preferences, we can leverage the larger model's generalizability and extensive knowledge base for superior performance across diverse tasks.

\subsection{Model-Specific Semantic Understanding}

Our findings suggest that models within the same family (e.g., Llama) exhibit a higher degree of semantic alignment compared to models from different families (e.g., GPT-4). Specifically, we observed that Llama-3 70B demonstrates a better understanding of rules generated by itself or the smaller Llama-3 8B model, compared to rules generated by GPT-4. While GPT-4 generated well-structured and seemingly comprehensive rules, they were less effective in guiding the Llama models. This indicates that semantic understanding, even when expressed through seemingly universal natural language, can be model-specific. 

This observation is further supported by experiments with human-written rules. Despite being crafted by expert annotators to be as clear and specific as possible, human-generated rules led to a 16.8\% performance degradation compared to model-generated rules. This suggests that subtle differences in the way models and humans interpret language can significantly impact the effectiveness of rule-based guidance. For instance, models might interpret terms like "precise," "concise," and "informal" differently than humans, leading to discrepancies between intended and actual outcomes.

These findings highlight the potential importance of model-specific semantic understanding in aligning LLMs with human preferences. Automated rule generation, leveraging the model's own internal representations and semantic understanding, is a more effective approach than relying on human-generated rules or prompts. However, further research is needed to fully understand the nature of these semantic differences and develop strategies for mitigating their impact.

\subsection{Enhancing Rule Generation through Deliberative Prompts}
\label{sec:discussion/thinking-tokens}

Humans often engage in a process of deliberation before formulating responses, particularly when faced with complex or nuanced tasks. This internal dialogue, where we weigh different options and consider various perspectives, contributes to more thoughtful and well-reasoned answers. Drawing inspiration from this human cognitive process, we explored the use of "deliberation" during rule generation inference by the preference agent. We include specially designed "thinking tokens," which encourage the model to engage in a similar form of internal reasoning before generating the natural language rules that guide the larger LLM. This encourages the model to decompose the task of preference extraction into smaller, more manageable steps. Our empirical results demonstrate that incorporating these deliberative prompts leads to a notable improvement in the quality of generated rules, resulting in better alignment between the large LLM's outputs and individual user preferences.

We hypothesize that these thinking tokens function as a form of cognitive scaffolding, providing the model with a structured space to isolate and process critical preference information. By explicitly prompting the model to "think" before generating rules, we aim to enhance its ability to identify subtle patterns in user preferences and translate them into effective guidance for the larger model. This approach aligns with findings from previous research, which demonstrates that prompting LLMs to engage in step-by-step reasoning can significantly improve their performance on various tasks \cite{kojima2023large, zelikman2024quietstar, goyal2024think}.  

\subsection{Evidence of Personalization}

A key objective of our approach is to learn individual writing styles rather than merely improving general task performance (e.g., email writing). To investigate this, we conducted a permutation analysis using preference agents trained on distinct email senders from the Enron dataset. We trained five agents on data from five different senders and then applied each agent to the test data of all five senders, generating emails for every agent-sender combination. This allowed us to assess whether an agent trained on a specific sender's style is more effective at generating emails resembling that sender's writing compared to other senders.

\begin{figure}
    \centering
    \includegraphics[width=\linewidth]{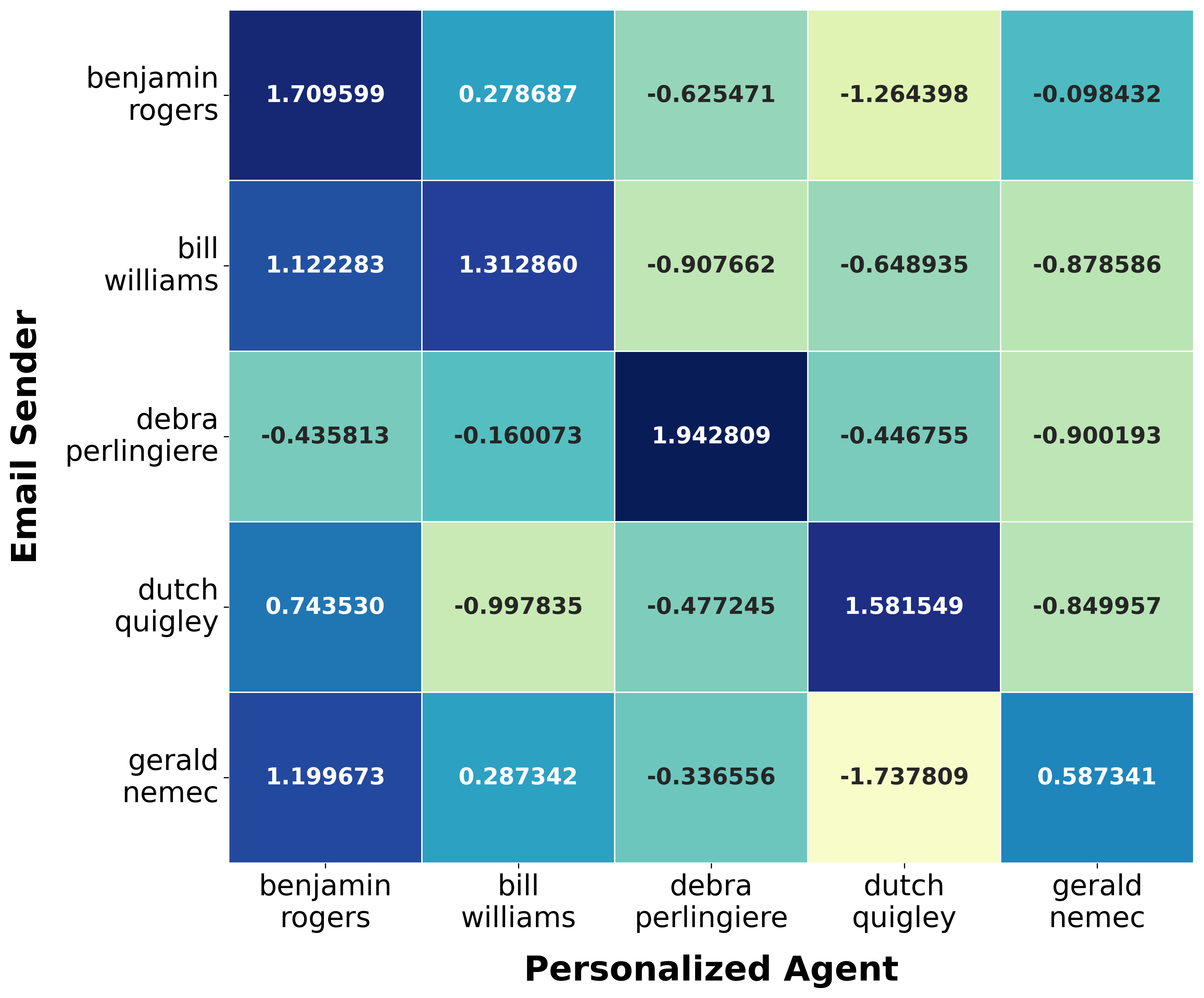}
    \caption{Permutation of Models and Senders}
    \label{fig:permutation}
\end{figure}

We quantified the similarity between the generated emails and the ground truth using the normalized BERT Score \cite{reimers2019sentencebert}, which provides a measure of semantic similarity suitable for analyzing large text corpora like emails. Our analysis, depicted in Figure \ref{fig:permutation}, reveals a strong trend along the diagonal. This indicates that the agent trained on a particular sender's data performs best when generating emails for that same sender, strongly suggesting that our approach successfully captures individual writing styles and preferences. 

This observation is further supported by randomly sampled human evaluations, which corroborate the BERT Score findings (see Appendix \ref{appendix:personalization} for details).

\subsection{Cost Effectiveness}
\label{discussion:cost-effectiveness}

While our approach necessitates an inference step with \(M_L\) during rule generation and at inference time, the cost, \(C_i(M_L)\), is relatively small due to the concise nature of the rule sequences. For instance, most rule sequences generated range from 100 - 150 extra tokens. This results in a combined cost of \(C_f(M_S) + C_i(M_L)\). Although this combined cost is marginally higher than the cost of naive fine-tuning (\(C_f(M_F)\)), the significant performance gains offered by our method, as evidenced by our experimental findings, justify this trade-off. Moreover, the inference cost associated with rule generation is a one-time expense during training data preparation, further diminishing its impact.

Our decision to avoid fine-tuning \(M_L\) provides significant flexibility as we avoid the sunk cost associated with fine-tuning a large model, enabling seamless integration of newer, more powerful LLMs as they become available.
\section{Related Work}

\textbf{Traditional Methods of Alignment.} Aligning language models to human preferences often employs techniques like Reinforcement Learning from Human Feedback (RLHF) \cite{ouyang2022training} and its variant, Reinforcement Learning from AI Feedback (RLAIF) \cite{bai2022constitutional}, which leverages fine-tuned LLMs as annotators. While effective, RLHF requires substantial human annotation and complex distributed training. Direct Preference Optimization (DPO) \cite{rafailov2023direct} offers an alternative by using preference pairs, reducing computational complexity. However, DPO's reliance on contrasting pairs may not fully capture the nuances of overlapping human preferences. In-context learning methods \cite{NEURIPS2022_8bb0d291, woźniak2024personalized}, while showing promise, are limited by context length restrictions, hindering their ability to generalize effectively.

\textbf{Agent-based Alignment.} To address the computational demands of training large models, agent-based architectures have emerged as a promising avenue for compute-constrained environments. For instance, \citet{li2023guiding} utilize a fine-tuned T5 policy model to guide large models via stimulus prompting. However, this approach necessitates full-parameter Supervised Fine-Tuning (SFT) and RL optimization, introducing computational overhead and yielding limited performance gains in tasks like dialogue generation. Similarly, Aligner \cite{ji2024aligner} employs full-parameter SFT and relies on a substantial custom dataset for preference learning, posing challenges in terms of data requirements and VRAM usage. \citet{tan2024democratizing} propose Parameter-Efficient Fine-Tuning (PEFT) methods to personalize agents based on user history and preference retrieval. While computationally efficient, this approach is constrained by the reasoning capabilities of the smaller fine-tuned agent. These approaches often rely on automatic metrics like BLEU and ROUGE, which predominantly capture lexical similarity without fully encapsulating the nuances of human preferences. \citet{gao2024aligning} introduce an agent trained on human edits to align zero-shot outputs. However, this approach requires multiple inference rounds for each query, increasing latency and computational costs. Moreover, human edit history may not consistently reflect genuine preferences, and relying solely on edit distance as a measure of alignment can be unreliable. \citet{yang2024aligning} propose a framework for aligning LLMs through Multi-perspective User Preference Ranking-based Feedback. This approach, however, involves an initial SFT phase, along with Multi-Perspective Ranking Aggregation (MPRA) and Reward Imitation Learning (RIL), leading to significant training overhead and the use of metrics like BLEU that may not accurately capture human preferences.

\textbf{Comparison with Aligner.} While both Aligner \cite{ji2024aligner} and our method utilize a small model trained with full-parameter SFT, our approaches differ significantly. Aligner focuses on correcting model outputs post-generation, while our preference agent proactively generates rules to steer the large model's initial output. This allows us to leverage the large model's reasoning capabilities by providing preference information upfront, rather than correcting its output afterwards. While Aligner demonstrates strong performance on tasks like text summarization and dialogue generation, its design is geared towards making smaller adjustments to large model outputs. Our task, on the other hand, often requires more substantial changes to align with user preferences, potentially necessitating complete rewrites of emails or articles. An Aligner-style approach or naive fine-tuning would face challenges in our setting, as a small model might struggle to accurately make drastic changes while preserving the large model's knowledge and coherence. This would also complicate the fine-tuning objective, as the patterns to be learned would be less well-defined and vary significantly across examples.

\section{Conclusion}

This work introduces a novel paradigm for aligning large language models (LLMs) with individual user preferences using limited data. We leverage small, locally trainable "preference agents" to efficiently guide larger LLMs without resource-intensive fine-tuning. Our approach generates natural language rules that encapsulate user preferences, acting as a "steering wheel" to direct the larger model's output towards desired styles and content.

This framework introduces a new preference fine-tuning objective: learning from implicit preference information found in the differences between a baseline LLM output and the user's desired output. This allows the agent to distill user preferences into actionable rules, enabling efficient personalization without modifying the larger model's weights.

Our empirical findings across diverse datasets demonstrate that preference agents significantly improve alignment with user preferences compared to existing methods in a compute-efficient manner, highlighting the potential for building highly personalized LLM applications at scale.
\clearpage
\section*{Limitations}

While our proposed method demonstrates significant improvements, there are a few areas for potential refinement. One consideration is the time required for the large model to process the preference agent's output before the first token can be generated. This could lead to a slightly higher Time to First Token (TTFT) at inference time. However, we believe the substantial performance gains offered by our approach outweigh this trade-off. 

As discussed in \S\ref{appendix:rulegen-strategies}, our most performant rule generation strategy incurs an additional computational cost compared to the alternative methods due to an extra zero-shot inference step. This cost is offset by the superior performance it enables.  We also provide a highly competitive "no-baseline" rule generation method which offers good performance at a lower inference cost. 

Furthermore, our rule generation strategy leverages thinking tokens, which can lead to slightly longer outputs. If output length is a strict constraint, this step can be omitted with minimal impact on the framework's effectiveness. Importantly, the inference cost associated with rule generation is a one-time expense incurred during training data preparation. 

Finally, as noted in \S\ref{discussion:cost-effectiveness}, using $M_L$ for preference agent rule generation introduces an additional inference iteration compared to naive fine-tuning.

While our current research focuses on text-based preferences, future work could explore extending this approach to other modalities, such as image or audio generation.  Additionally, investigating the integration of multimodal preferences and the development of more sophisticated rule generation strategies could further enhance the capabilities of preference agents. We believe that this research opens exciting new avenues for personalized LLM applications, paving the way for a future where powerful language models can be seamlessly tailored to individual needs and preferences, ultimately enhancing user experience and fostering more engaging and human-centric interactions. 
\section*{Ethical Considerations}

In this work, we have taken several steps to ensure that our research adheres to ethical principles and respects the rights of all parties involved. We are committed to the responsible and ethical use of AI technology and have implemented measures to prevent potential misuse of our work.

\paragraph{Dataset Licensing and Attribution.} Both datasets used in this research will be released under the Creative Commons Attribution-NonCommercial 4.0 International (CC BY-NC 4.0) license.

The Enron email dataset \cite{10.1007/978-3-540-30115-8_22} is available for educational and research purposes under the principles of fair use. We have credited the original dataset creators and adhered to the terms of its usage.

The New Yorker dataset is based on the 'All the News 2.0' dataset by Andrew Thompson \cite{allthenews2}, which is licensed for non-commercial, research purposes only. We have made modifications and enhancements to the dataset, and these changes are also licensed under the CC BY-NC 4.0 license. We have properly attributed the original dataset and its creator.

\paragraph{Model Release.} In compliance with the terms of the 'All the News 2.0' dataset license, we will not be releasing the fine-tuned agents trained on the New Yorker dataset. The license explicitly states that the dataset is to be used for research purposes only and not for the release of commercial generative models.

Similarly, we will not release the agent fine-tuned on the Enron email corpus. This decision was made to ensure that our models are not used to impersonate the senders in the Enron email corpus without their explicit permission. We believe that releasing such a model could potentially infringe upon the privacy rights of the individuals involved. 

However, for research purposes only, we will make the models available upon request.

\paragraph{Citation and Acknowledgment.} We have taken extensive care to ensure that we comply with all licenses and have appropriately cited any of our work that is a derivative of another project. We acknowledge the original creators and their contributions to the field.

\paragraph{Potential Misuse.} We acknowledge that our datasets, though open-source, can potentially be used to train AI assistants or models for malicious purposes. We strongly condemn any misuse of our work and explicitly support the safe and responsible use of AI technology. Our intention is to advance the field of AI research while adhering to ethical principles and preventing harm.

\section*{Acknowledgements}

We would like to express our sincere gratitude to the reviewers of ACL ARR for their insightful and constructive feedback, which significantly improved the quality of this paper. We are particularly grateful for their suggestion to incorporate the LAMP dataset, which broadened the scope and impact of our evaluation.
\bibliography{resources/main}

\clearpage
\appendix
\section{Datasets Overview}
\label{appendix:dataset-enron}

For the Enron dataset, we began with the original Enron email corpus. To focus on original content creation, emails containing only forwarded content like email threads, blog posts, and articles were removed. We then dissected the remaining emails into two distinct parts: \textbf{previous\_context} encompassing any preceding email chain or reply content, and \textbf{content} representing the original message drafted by the sender. This careful separation, achieved through a specifically designed heuristic, ensured that \textbf{only self-written} content was considered during analysis. After these steps, we release our dataset - Enron-42k.

Conversely, the New Yorker dataset required minimal pre-processing. This dataset, comprising articles from the New Yorker publishing house, was already cleaned, pre-processed, and structured with the necessary features for our study. As such, we utilized the New Yorker dataset in its original form.

The LAMP 3U Amazon reviews dataset consists of customer reviews for a specific product. We selected this dataset to explore the application of our methods in a product review domain. Similar to the Enron and New Yorker datasets, the goal was to leverage user-generated content to understand preferences and generate tailored responses. We extracted user intents from the reviews and used these intents to create baselines and rules for fine-tuning our preference agents. This approach mirrors the methodology applied to the other two datasets, allowing for a consistent evaluation framework across different domains.

\begin{figure}[H]
    \centering
    \includegraphics[width=\linewidth]{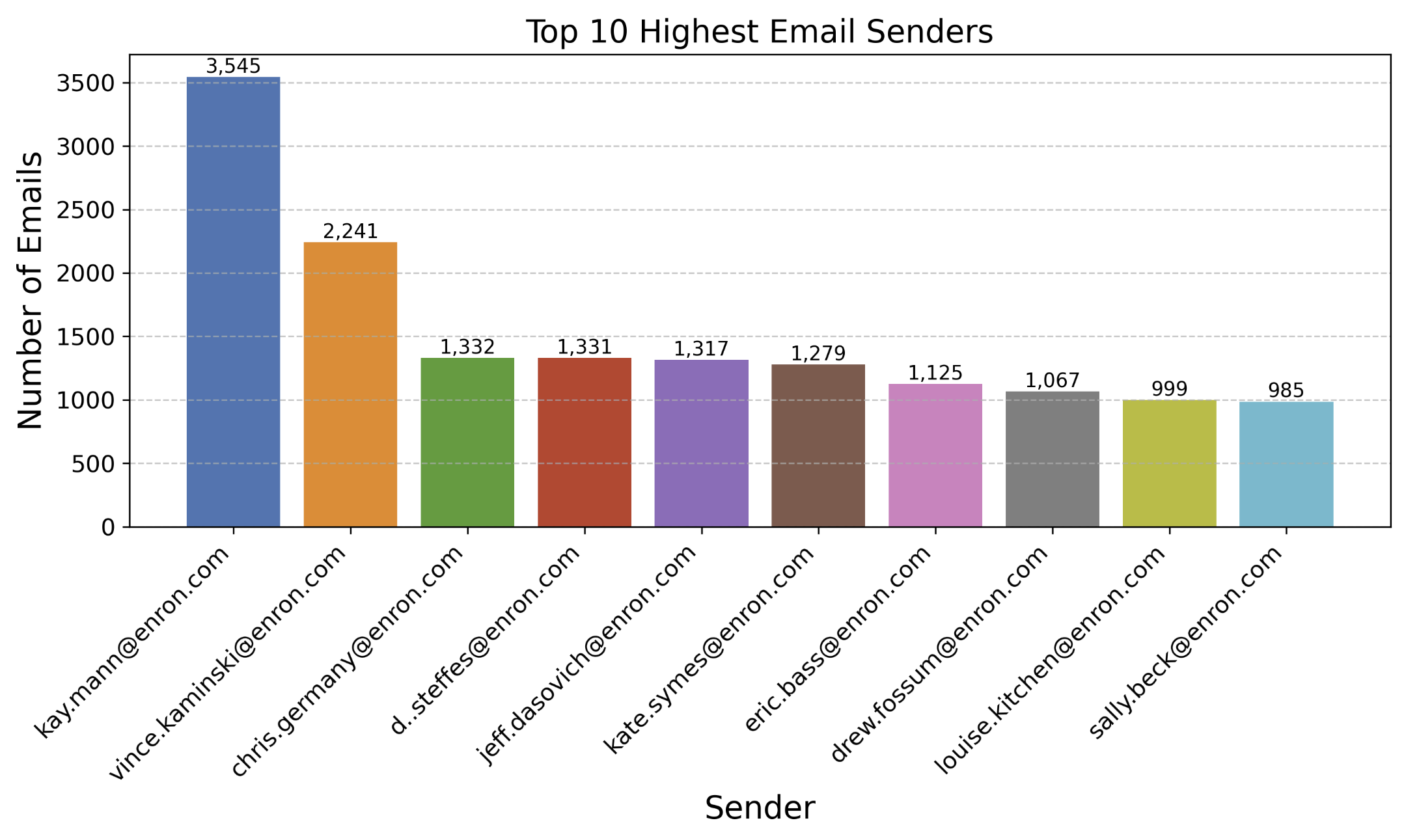}
    \caption{Top 10 senders for the Enron-42k Dataset}
    \label{fig:enron-senders}
\end{figure}

\begin{figure}[H]
    \centering
    \includegraphics[width=\linewidth]{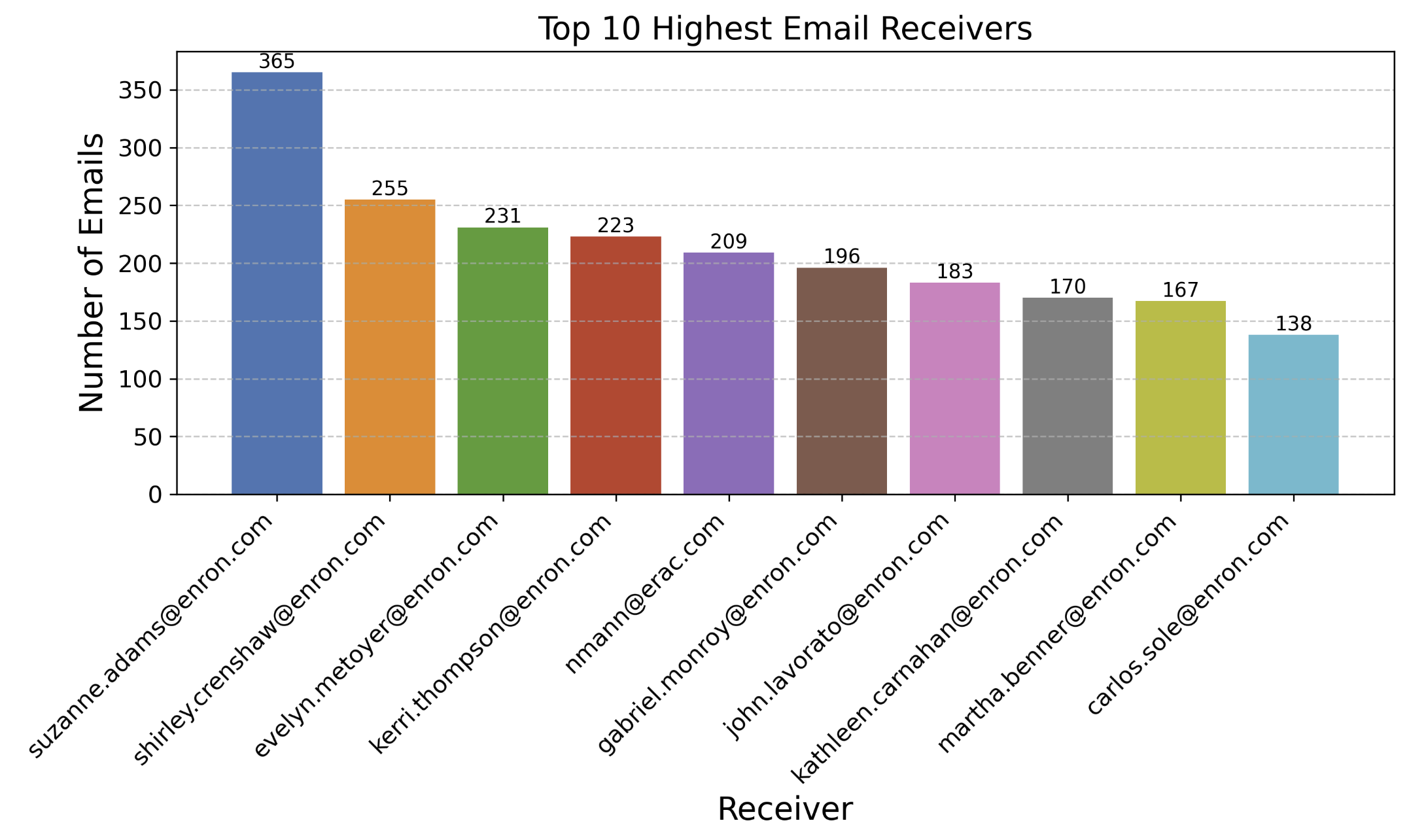}
    \caption{Top 10 receivers for the Enron-42k Dataset}
    \label{fig:enron-receivers}
\end{figure}

\begin{figure}[H]
    \centering
    \includegraphics[width=\linewidth]{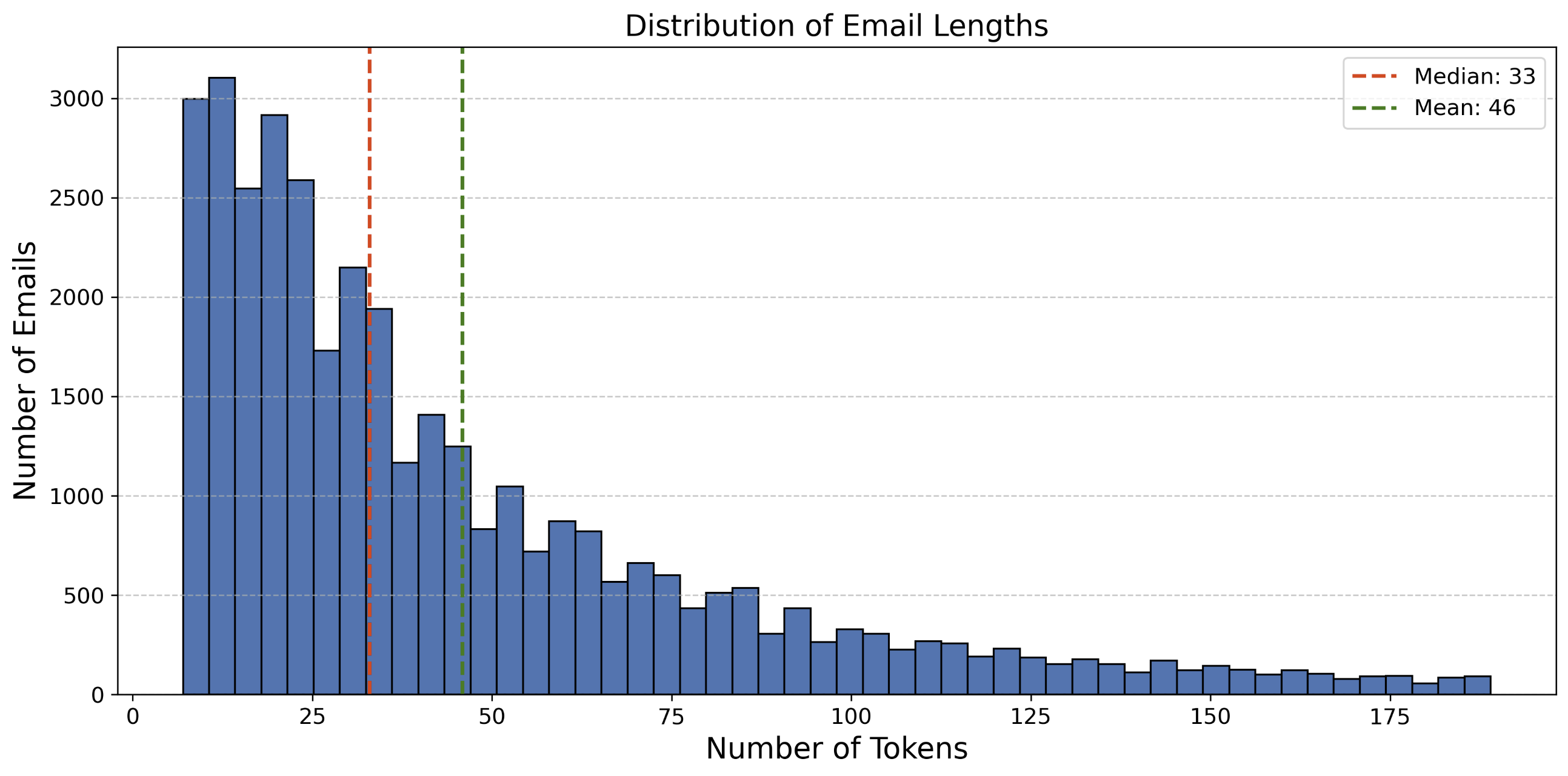}
    \caption{Distribution of token lengths for emails in the Enron Dataset}
    \label{fig:enron-distribution}
\end{figure}

\begin{figure}[H]
    \centering
    \includegraphics[width=\linewidth]{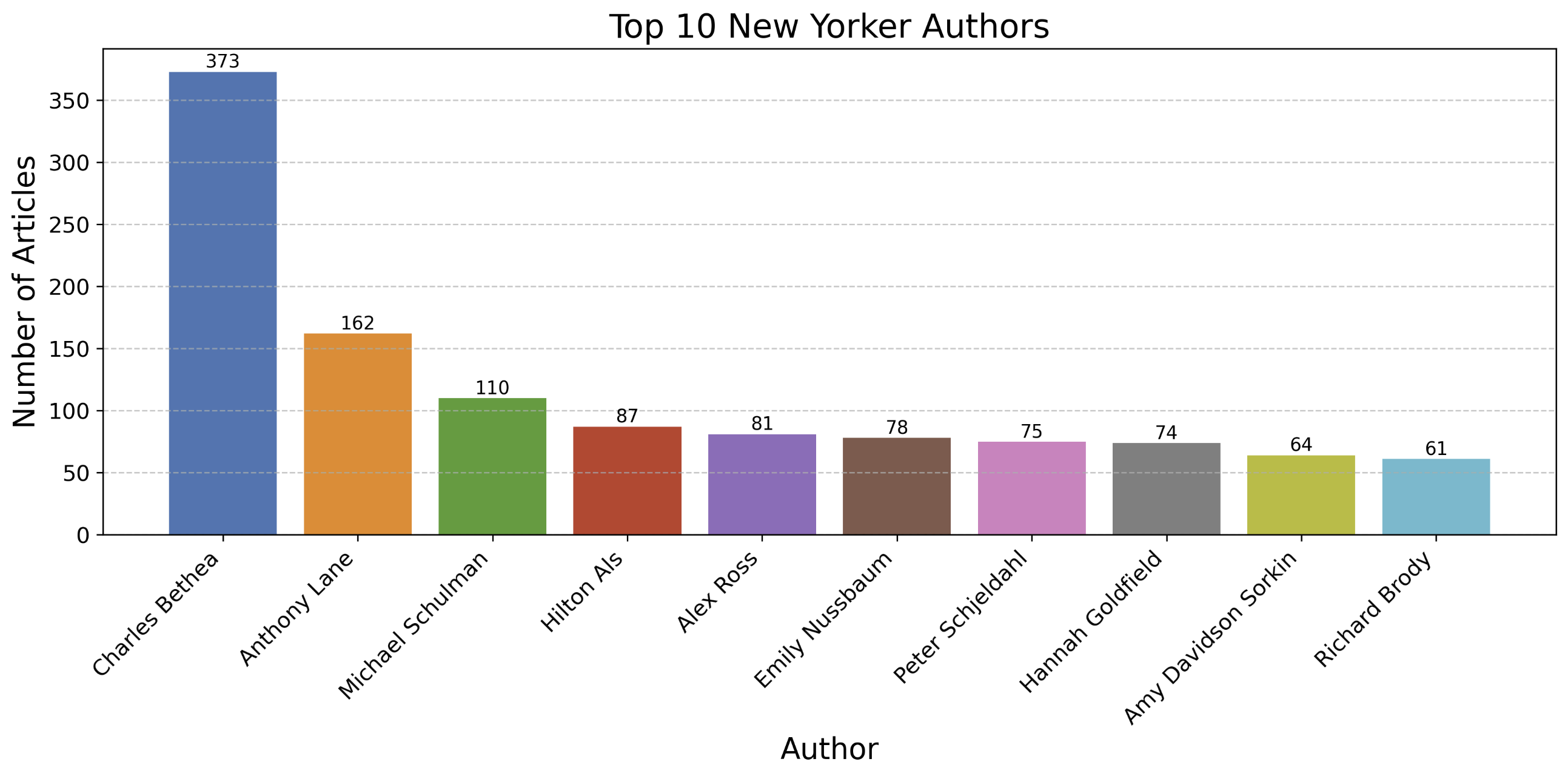}
    \caption{Top 10 authors in the New Yorker Dataset}
    \label{fig:ny-authors}
\end{figure}

\begin{figure}[H]
    \centering
    \includegraphics[width=\linewidth]{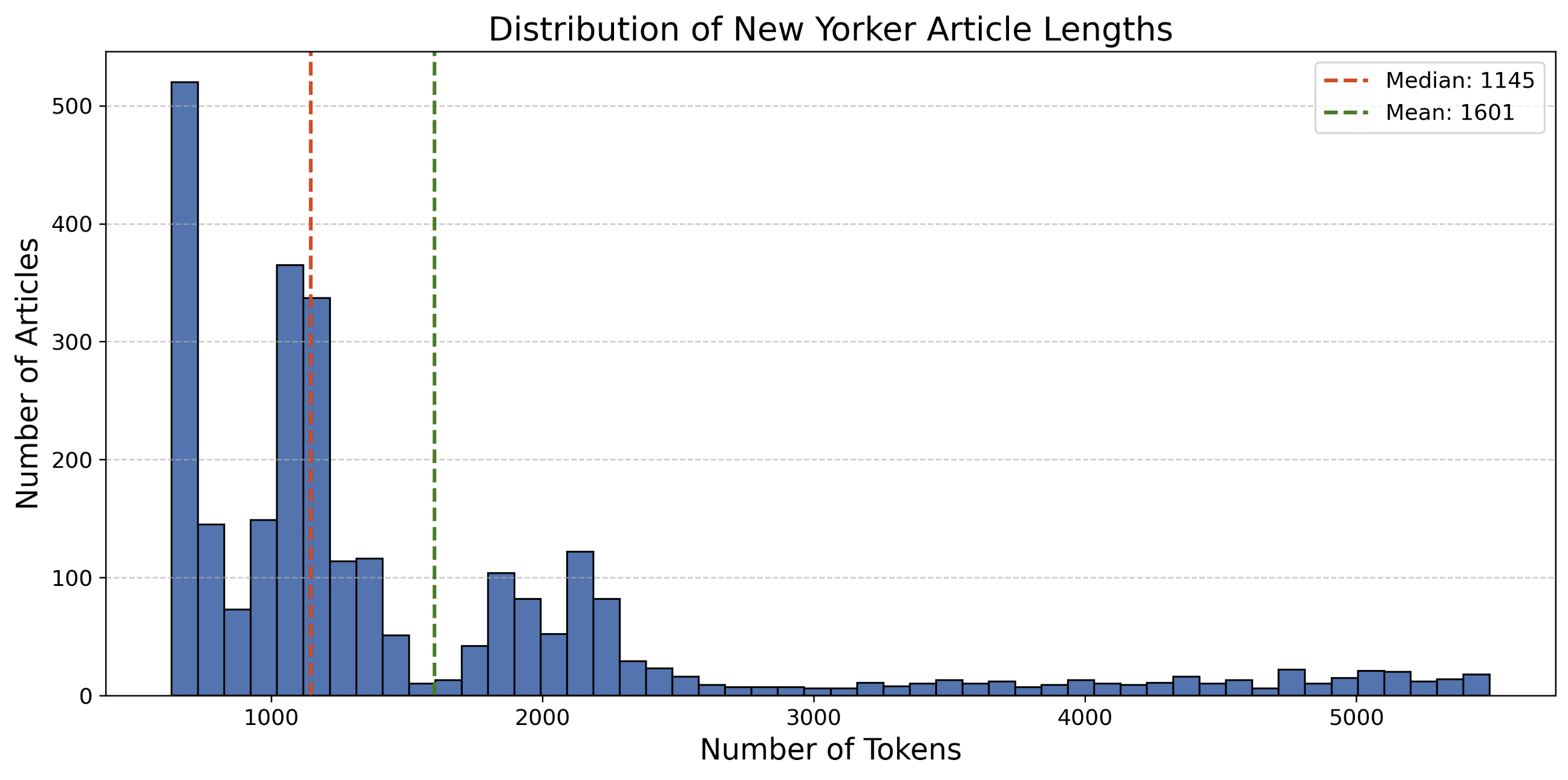}
    \caption{Distribution of token lengths for articles in the New Yorker Dataset}
    \label{fig:ny-distribution}
\end{figure}

\begin{figure}[H]
    \centering
    \includegraphics[width=\linewidth]{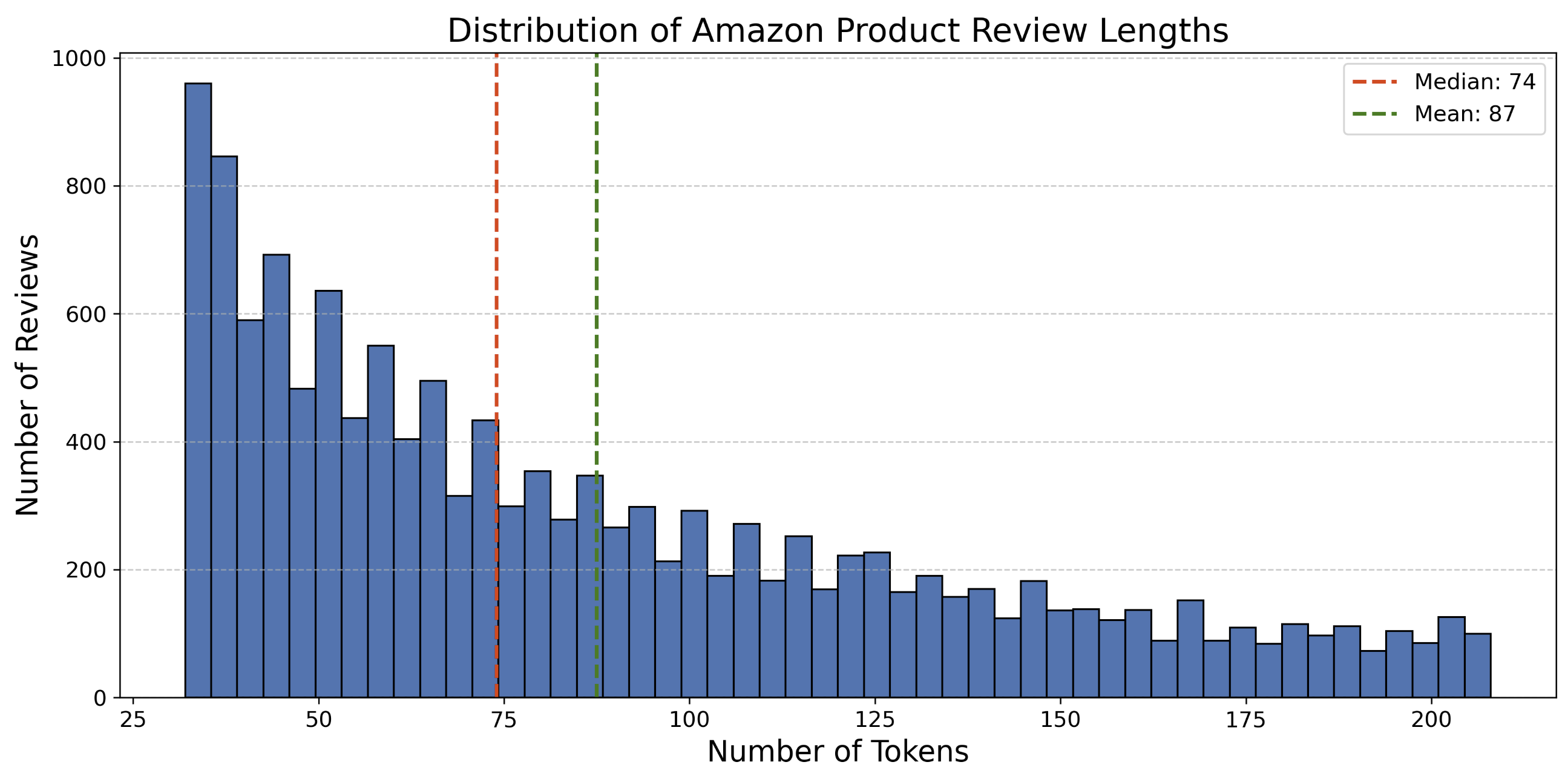}
    \caption{Distribution of token lengths for reviews in the LAMP 3U Dataset}
    \label{fig:lamp-distribution}
\end{figure}

\section{Generations}

\subsection{Generated Rules}
\label{appendix:generations/rules}

The rules were generated using preference agents for the following ground truth email:

\begin{lstlisting}
Paul, Here is an updated version of the agreement I sent to Steve on Friday. Sorry I didn't cc you to start with. It's pretty much exactly as we discussed. Let me know if you have any questions. Thanks, Bill
\end{lstlisting}

Are:

\begin{lstlisting}
1. **Tone**: Use a conversational tone and language throughout the email. 2. **Length**: Keep the email concise, aiming for a total of around 50-60 words. 3. **Introduction**: Begin the email with a brief introduction that sets the context, using a phrase like "Here is an updated version of the agreement". 4. **Specific details**: If applicable, mention any relevant background information, such as who else the agreement was sent to and when. 5. **Apology phrasing**: Use a concise apology phrase like "Sorry I didn't cc you to start with" instead of a longer, more formal apology. 6. **Request for feedback**: Use a brief, informal phrase like "Let me know if you have any questions" to request feedback or questions from the recipient. 7. **Closing sentence**: End the email with a concise, informal phrase like "Thanks" instead of a more formal closing like "Best". 8. **Signature**: Use a simple, formal signature with the sender's name. 9. **Sentence structure**: Use short, simple sentences throughout the email to maintain a conversational tone. 10. **Language**: Avoid using overly formal or technical language, opting instead for a more conversational, everyday tone.
\end{lstlisting}

\subsection{Thinking Tokens}
\label{appendix:thinking_tokens}
\begin{lstlisting}
<thinking>
The ground truth email is a concise and informative response to the previous context, discussing market trends and trading activities. Key observations include:
1. **Length and Structure**: The email is relatively short, consisting of 8 sentences. It uses a straightforward structure, jumping directly into the observations and insights without any formal greetings or sign-offs.
2. **Tone and Style**: The tone is informal and conversational, with a focus on market terminology and shorthand that would be understood by someone familiar with the trading environment. There are contractions (e.g., "saw", "taking off"), colloquial phrases (e.g., "get whacked"), and specific market terms (e.g., "VAR", "h/j spread").
3. **Common Phrases and Terminology**: Specific terms like "bulls", "summer against length", "mitigate margins", "absolute position limits", "VAR", "Access", and "bearish to neutral trade" are used. These reflect the insider knowledge and jargon of the trading community.
4. **Observational Detail**: The email provides specific details about market activities and price changes. Mention key players (e.g., El Paso) and their actions in the market.
5. **Consistency with Intent**: The content matches the intent closely, summarizing market movements, trading strategies, and potential future movements. Reflect the key points mentioned in the intent, such as bulls' behavior, El Paso's buying, and the change in the H 20 call price.
6. **Grammar and Punctuation**: The email uses casual grammar and minimal punctuation. Avoid unnecessary capitalization and formal sentence structures.
</thinking>
<rules>
1. **Length and Structure**: The email should be concise, consisting of approximately 8 sentences. Avoid adding formal greetings or sign-offs.
2. **Tone and Style**: Use an informal and conversational tone. Include contractions and colloquial phrases to reflect the casual nature of the communication.
3. **Market Terminology**: Incorporate specific market-related terms and shorthand (e.g., "VAR", "h/j spread", "Access", "bearish to neutral trade"). Ensure the language reflects insider knowledge and jargon of the trading community.
4. **Observational Detail**: Provide specific details about market activities and price changes. Mention key players (e.g., El Paso) and their actions in the market.
5. **Consistency with Intent**: Ensure the content matches the intent closely, summarizing market movements, trading strategies, and potential future movements.
6. **Grammar and Punctuation**: Use casual grammar and minimal punctuation. Avoid unnecessary capitalization and formal sentence structures.
7. **Specific Phrases**: Use phrases like "saw a lot of the bulls", "mitigate margins/absolute position limits/var", "taking off spreads", "trading equivalent of 180 on access", "get whacked", "if h settles \$20", "H 20 call was trading for 55 on monday", "market's view of probability of h going crazy".
8. **Flow and Coherence**: Ensure the email flows logically from one point to the next, maintaining coherence while jumping between observations and insights.
</rules>
\end{lstlisting}

\subsection{Generated Intents}
\label{appendix:generated_intents}

Here is an example generated intent of a casual work conversation:
\begin{lstlisting}
Original Email and Metadata:
---
From: jeff.dasovich@enron.com
To: michael.etringer@enron.com
Date: Mon, 29 Oct 2001 13:28:55 -0800 (PST)
Subject: [None]

Previous Context:
[None]


Email:
You won't believe this. 

Jerry Bloom left me a message. Wants to talk, though I wouldn't get my hopes up. 

I'll let you know what he says. 

Best, 
Jeff
---

Generated Intents:
---
Variant 1:
<thinking>
1. Identify the main event mentioned in the email.
2. Note the person involved and their action.
3. Mention Jeff's skepticism about the conversation.
4. Highlight the intention to keep Michael informed.
5. Ensure the bullet points are concise and focused on the key information.
</thinking>
<bullet_points>
- Received a message from Jerry Bloom wanting to talk
- Skeptical about the conversation
- Will update Michael on what Jerry says
- Will keep Michael informed
</bullet_points>
---
Variant 2:
<thinking>
1. Identify the main surprise or unexpected event mentioned in the email.
2. Note who initiated the contact and what they want to do.
3. Capture Jeff's skepticism and uncertainty about the outcome.
4. Mention Jeff's intention to follow up with Michael.
</thinking>
<bullet_points>
- Jerry Bloom left a message for me, wants to talk
- I'm skeptical and wouldn't get hopes up
- Will let Michael know what Jerry says
- Uncertainty about the outcome
</bullet_points>
---
\end{lstlisting}
\section{Rule Generation Strategies and Their Implications}
\label{appendix:rulegen-strategies}

We explored three distinct strategies for rule generation, each offering insights into how best to capture and leverage user preferences for guiding large language models.

\noindent \textbf{$R_1$} involved prompting the large language model ($M_L$) to generate rules based on user intent and task metadata. While this approach provides a basic framework for rule generation, it lacks the depth of analysis needed to accurately distill user preferences. 

\noindent \textbf{$R_2$} extended $R_1$ by incorporating "thinking tokens," (see \S\ref{sec:discussion/thinking-tokens}), prompting $M_L$ to engage in a more deliberate reasoning process before generating rules. This resulted in modest improvements, highlighting the benefits of encouraging more thoughtful rule generation.

\noindent \textbf{$R_3$}, based on a distillation process, proved most effective. This strategy leverages $M_L$'s zero-shot output as a starting point, prompting it to identify discrepancies between its initial response and the ground truth. By explicitly focusing on these differences, $M_L$ generates rules specifically designed to address the missing preference information. This targeted approach led to significant performance gains, with $R_3$ outperforming both $R_2$ and $R_1$ by 65\% on the Enron dataset and 69.7\% on the New Yorker dataset.

By explicitly identifying the gaps in preference alignment, the distillation process enables the generation of highly targeted and effective rules.

While $R_3$ offers the best performance, it is worth noting that it incurs an additional inference cost compared to $R_1$ and $R_2$ due to the extra zero-shot generation step. In scenarios where computational resources are limited, $R_2$, which leverages thinking tokens for improved rule generation without the added inference cost, provides a compelling alternative.

\section{Finetuning Hyperparameter Search For Rule Generators}
\label{appendix:finetuning-analysis}

To identify the optimal configuration, we train four rule generators on our gold-standard rules, varying the ranks in each case. We implement a 1:1 mapping between the LoRA rank and Alpha.

\begin{figure}[bt!]
    \centering
    \includegraphics[width=\linewidth]{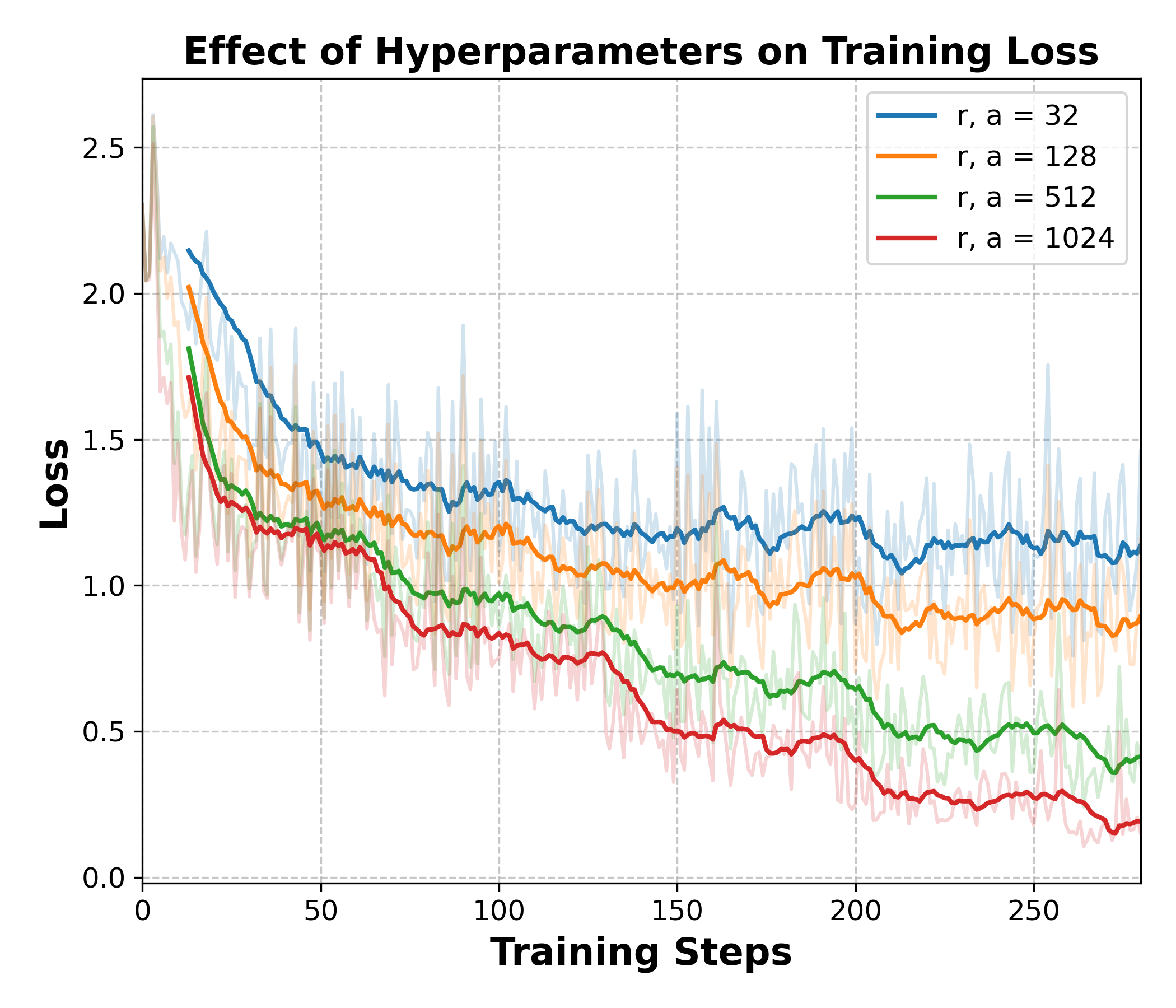}
    \caption{Rule Generator Hyperparameter Search}
    \label{fig:rule-gen-training-runs}
\end{figure}

As anticipated, our results indicate that higher Alpha values and corresponding ranks lead to improved training losses. This trend is illustrated in Figure \ref{fig:rule-gen-training-runs}, which shows the relationship between increasing Alpha/rank values and the resulting training performance. These findings underscore the importance of selecting appropriate parameter settings to optimize the rule generator's effectiveness.

\section{Similarity Metrics}
\label{appendix:automated-metrics}

This work evaluates the similarity between responses generated by different methods and the ground truth for a given task. Our primary goal is to assess how effectively each method captures the user's preferences in terms of style, tone, and word choice. While metrics like BLEU, ROUGE, and TFIDF Cosine similarity are commonly used to evaluate the lexical overlap between texts, they fall short of capturing the nuanced aspects of stylistic similarity crucial to our evaluation.

\begin{figure}[b]
    \centering
    \includegraphics[width=0.9\linewidth]{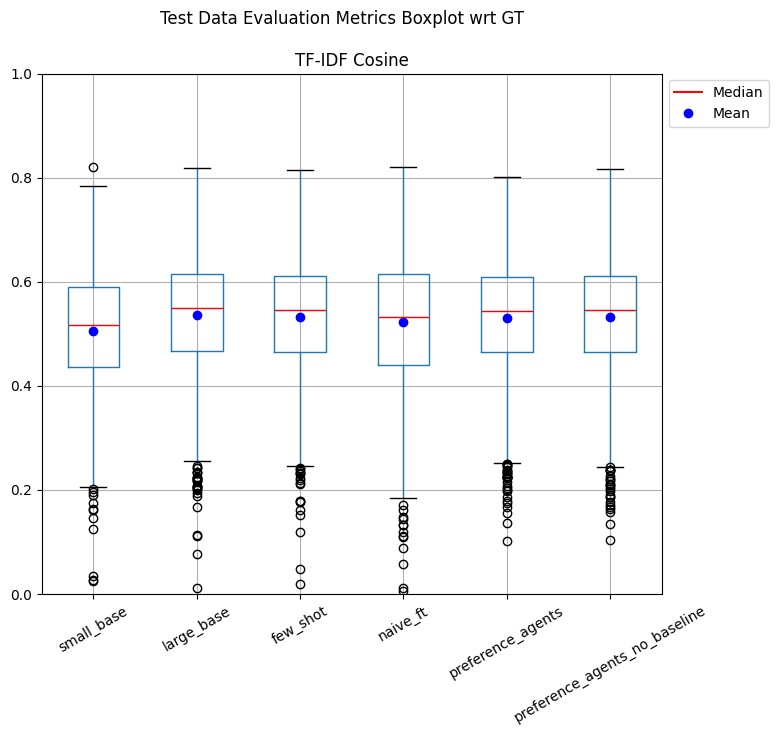}
    \caption{New Yorker TF-IDF Similarity Scores}
    \label{fig:similarity-metrics}
\end{figure}

Consider the example in Figure \ref{fig:similarity-metrics}, which shows the TF-IDF cosine similarity scores for different methods on the New Yorker dataset. While there are slight variations in the median scores, the overall distributions largely overlap. This suggests that TF-IDF, which primarily relies on term frequency and inverse document frequency, struggles to differentiate between outputs that exhibit clear stylistic differences to human readers or as evaluated by GPT-4o. Similarly, in the Enron dataset, an email using formal language like "Dear Mr. Smith" and "Sincerely" might receive a similar BLEU score to an email using informal greetings like "Hey John" and "Cheers" despite the contrasting styles. This highlights the limitations of BLEU in capturing the subtle variations in word choice that contribute to a specific writing style.

Several alternative metrics have been proposed to address the shortcomings of traditional lexical overlap measures. BERT Score \cite{reimers2019sentencebert}, for instance, leverages pre-trained BERT embeddings to compute semantic similarity between sentences, potentially capturing stylistic nuances better than BLEU or ROUGE. Similarly, BLEURT \cite{sellam-etal-2020-bleurt} is a learned metric that utilizes a large pre-trained language model to predict human judgments of translation quality, which can be adapted to assess stylistic similarity. However, even these advanced metrics might not fully capture the complexities of human preferences for style and tone, which can be subjective and context-dependent \cite{Peng_2017}.

Given these limitations, we prioritized GPT-4o evaluation and human evaluation for our analysis. Human judgment remains the gold standard for evaluating stylistic similarity, as it reflects the inherent subjectivity of human preferences. GPT-4o, with its advanced language understanding capabilities, can serve as a reliable proxy for human judgment, particularly in capturing stylistic nuances \cite{naismith-etal-2023-automated}. By combining GPT-4o evaluation with a focused human evaluation study, we aim to provide a comprehensive and nuanced assessment of the alignment of model outputs with individual user preferences.
\section{Personalization Test}
\label{appendix:personalization}

While the diagonal trend generally holds, there are instances where an agent trained on one sender performs well across multiple senders. For example, the agent trained on Benjamin Rogers' emails exhibits high BERT Scores across all senders. We hypothesize that this may be due to the diversity of Rogers' email interactions and the larger size of his training set, which allows the model to learn the underlying task exceptionally well. Consequently, this agent demonstrates strong performance even when generating emails for other senders, highlighting the model's ability to generalize beyond individual preferences when trained on sufficiently diverse data.

Here are the un-normalized BERT Score values for the personalization test (for 5 Enron employees). Though these aren't a perfect metric, they provide a generalized view of the large evaluation space that we have:

\begin{table}[h!]
    \centering
    \resizebox{\linewidth}{!}{%
    \begin{tabular}{@{}lccccc@{}}
        \toprule
        & \textbf{Benjamin Rogers} & \textbf{Bill Williams} & \textbf{Debra Perlingiere} & \textbf{Dutch Quigley} & \textbf{Gerald Nemec} \\
        \midrule
        \textbf{Benjamin Rogers} & 0.907984 & 0.883311 & 0.867720 & 0.856703 & 0.876808 \\
        \midrule
        \textbf{Bill Williams} & 0.857471 & 0.858338 & 0.848238 & 0.849415 & 0.848370 \\
        \midrule
        \textbf{Debra Perlingiere} & 0.818253 & 0.821676 & 0.847782 & 0.818117 & 0.812488 \\
        \midrule
        \textbf{Dutch Quigley} & 0.809500 & 0.804509 & 0.806001 & 0.811901 & 0.804933 \\
        \midrule
        \textbf{Gerald Nemec} & 0.858304 & 0.852070 & 0.847807 & 0.838231 & 0.854120 \\
        \bottomrule
    \end{tabular}%
    }
    \caption{Bert Score Values for different individuals (unnormalized)}
    \label{tab:values}
\end{table}

\begin{figure}[h]
    \centering
    \includegraphics[width=\linewidth]{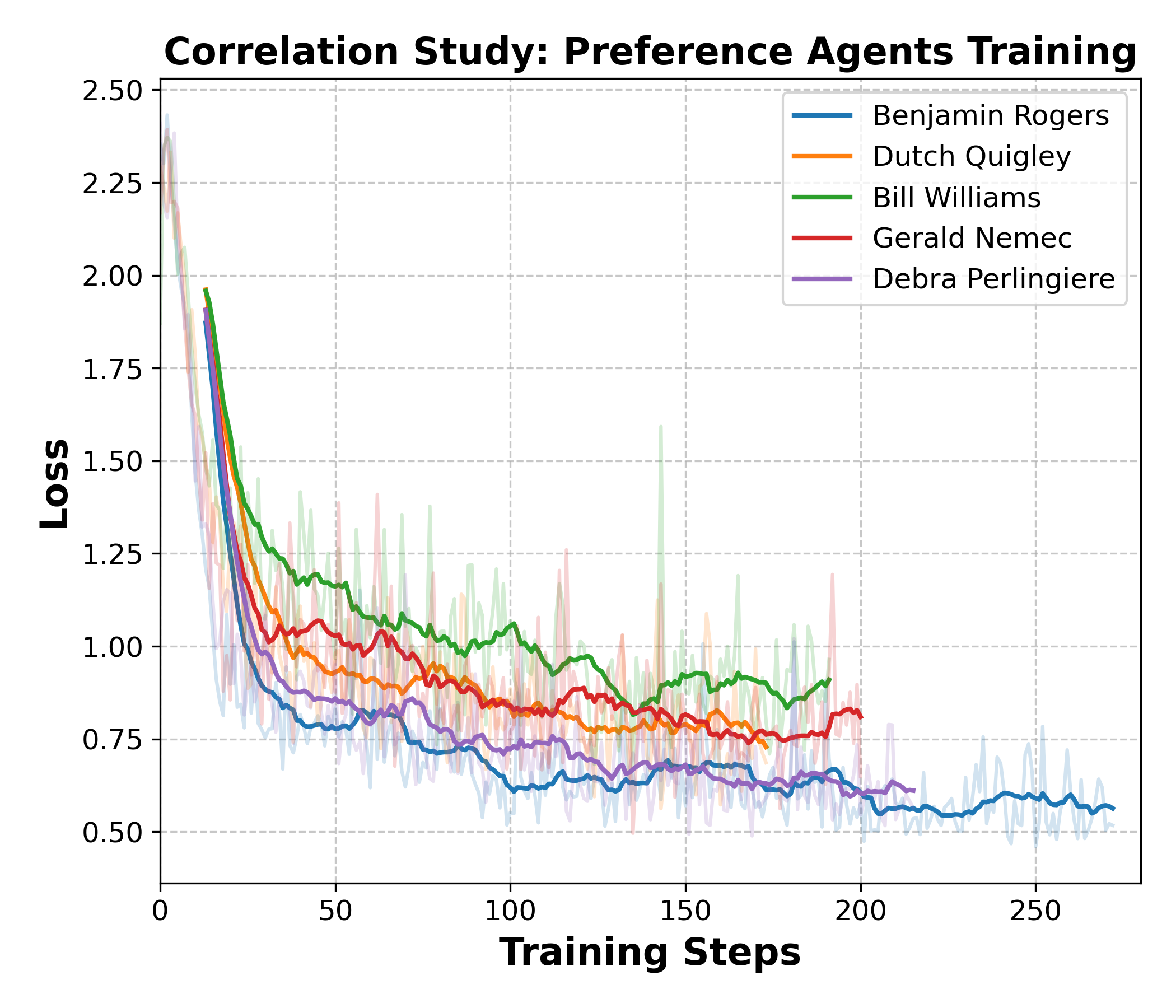}
    \caption{Train Loss For Preference Agents}
    \label{fig:train-loss}
\end{figure}                        
\section{Compute Infrastructure}
\label{appendix:compute-infra}

Experiments were run on NVIDIA 8xH100 nodes, for Llama 70B inference and generations. Finetuning was tested on both NVIDIA A5000 (to simulate consumer infrastructure) and NVIDIA A100 GPUs.
\section{Prompts}
\subsection{Intent Generation}
\label{appendix:prompts/intent-generation}
\subsubsection{Enron Intent Generation}

\begin{lstlisting}
You will be given an email and some surrounding context. Your task is to extract the core content of the email, omitting any stylistic or extraneous elements.

First, carefully read through the entire email and context. Then, reflect on the main purpose and key points of the email in a <scratchpad>. Consider what the sender is trying to communicate and what information is most essential.

<scratchpad>
<!-- Use this space to reflect on the main purpose and key points of the email. -->
</scratchpad>

Finally, extract the core content of the email in bullet point form. Omit any stylistic elements like greetings, sign-offs, pleasantries, etc. Focus solely on the key information and action items. Provide your extraction inside <core_content> tags. The core content, should be in first person format (for the email sender). Think and reflect extensively, to make sure you get the details right.

<core_content>
<!-- Extract the core content of the email here in bullet point form. -->
</core_content>
\end{lstlisting}

\subsubsection{New Yorker Intent Generation}

\begin{lstlisting}
You will be given a news article and some surrounding context. Your task is to extract the core content of the article, omitting any stylistic or extraneous elements.

First, carefully read through the entire article and context. Then, reflect on the main purpose and key points of the article in a <scratchpad>. Consider what the writer is trying to communicate and what information is most essential.

<scratchpad>
<!-- Use this space to reflect on the main purpose and key points of the article -->
</scratchpad>

Finally, extract the core content of the article in bullet point form. Omit any stylistic elements like tone, style, sign-offs, etc. Focus solely on the key information and action items. Provide your extraction inside <core_content> tags. Please include any direct quotes from the article in the core content. Write the core points from the writers perspective. Think and reflect extensively, to make sure you get all the details right.

<core_content>
<!-- Extract the core content of the article here in bullet point form. -->
</core_content>
\end{lstlisting}

\subsubsection{Lamp3U Review Intent Generation}
\begin{lstlisting}
You will be given a review and some surrounding context. Your task is to extract the core content of the review, omitting any stylistic or extraneous elements.
First, carefully read through the entire review and context. Then, reflect on the main purpose and key points of the review in a <scratchpad>. Consider what the reviewer is trying to communicate, what aspects of the item they're reviewing, and what information is most essential.

<scratchpad>
<!-- Use this space to reflect on the main purpose and key points of the review. -->
</scratchpad>

Finally, extract the core content of the review in bullet point form. Omit any stylistic elements like flowery language, personal anecdotes, or repetitive praise/criticism. Focus solely on the key information, opinions, and specific details about the reviewed item. Provide your extraction inside <core_content> tags. The core content should be in first person format. Include any notable quotes or specific examples given in the review. Think and reflect extensively to ensure you capture the essence of the review accurately.

<core_content>
<!-- Extract the core content of the review here in bullet point form. -->
</core_content>
\end{lstlisting}

\subsection{Rule Generation}

\subsubsection{Enron Email Dataset}

\textbf{No Baseline Email Rule Generator}
\begin{lstlisting}
You are an expert rule generator whose task is to generate a detailed set of rules given the metadata of an email, previous context, user intent, and the ground truth email. First you must go through the metadata carefully, analyzing who the sender and receiver is, the subject of the email, and the user intent. After analyzing this information, please generate a set of extremely detailed and granular set of rules that would help a model generate an email that is exactly how the user would intent to write it. Make sure the rules are specific to the given user and receiver pair and pay close attention to the user intent. Please generate these extremely detailed, specific, and granular set of rules.
\end{lstlisting}

\textbf{With Baseline Email Rule Generator}
\begin{lstlisting}
You are an expert rule generator whose task is to ensure that a base email can be transformed into the ground truth email. You are provided with the following: The intents that were used to generate the base email, the base email and the ground truth email. You must analyze the differences between the base email and the ground truth email in great detail analyzing every difference. You must focus on the following while generating these rules: the difference in the length of the emails, the tone, style, structure, common phrases, nicknames, signature, and anything else that you think is very important. All these factors must be closely analyzed to generate these extremely granular set of rules. Please also mention exactly how long the email should be and generate an extremely detailed and granular set of rules that should be able to transform the base email exactly into the ground truth email. To do this please first think deeply and analyze these differences within <thinking></thinking> tags where you can enlist every possible difference between the base and the ground truth email. Once this is done please generate an extremely detailed and granular set of rules that can be used to transform the base email. Do not mention the ground truth email in your set of rules whatsoever and do not talk about removing things from the base email. The rules should be an extremely detailed guideline to transform the base to ground truth email. The rules should not reference the ground truth or base email, and should be a standalone list of detailed rules. Please include these detailed set of rules within <rules></rules> tags.
\end{lstlisting}

\subsubsection{New Yorker Dataset}

\textbf{No Baseline Rule Generation}
\begin{lstlisting}
You are an expert rule generator whose task is to help a model generate articles that are close to the ground truth article given user intent. You are given some metadata and the user intent which is the input to generate an article, and the ground truth article. Your task is to deeply analyze the intents and ground truth very carefully and generate a set of rules that you think are very important to fully capture the nuances of the ground truth article. While analyzing the article please consider the following factors: the exact length of the article, the tone, writing style, structure, important phrases, direct quotes, and anything else that you think is very important. First start by analyzing the ground truth article extremely carefully accounting for all the important factors within <thinking></thinking> tokens. Once you have done that list a set of extremely detailed and granular rules to ensure that all nuances of the ground truth article are captured to ensure that the generated article is exactly the ground truth article. Include everything including phrases that are important and all stylistic information that needs to be captured in extreme detail. Please enclose these extremely detailed, specific, and granular set of rules within <rules></rules>
\end{lstlisting}

\textbf{With Baseline Rule Generations}

\begin{lstlisting}
You are an expert rule generator whose task is to ensure that a base article can be transformed into the ground truth article. You are provided with the following: The intents that were used to generate the base article, the base article and the ground truth article. You must analyze the differences between the base and the ground truth in great detail analyzing every difference. You must focus on the following while generating these rules: the difference in the length of the articles, the tone, style, structure, common phrases, nicknames, signature, and anything else that you think is very important. All these factors must be closely analyzed to generate these extremely granular set of rules. Please also mention exactly how long the article should be and generate an extremely detailed and granular set of rules that should be able to transform the base article exactly into the ground truth article. To do this please first think deeply and analyze these differences within <thinking></thinking> tags where you can enlist every possible difference between the base and the ground truth article. Once this is done please generate an extremely detailed and granular set of rules that can be used to transform the base article. Do not mention the ground truth or base article in your set of rules whatsoever. The rules should be an extremely detailed guideline to transform the base to ground truth article. Please include these detailed set of rules within <rules></rules> tags.
\end{lstlisting}

\subsubsection{Lamp3U Review Dataset}
\textbf{No Baseline Rule Generation}
\begin{lstlisting}
You are an expert rule generator whose task is to help a model generate reviews that are close to the ground truth review given user intent. You are given some metadata and the user intent which is the input to generate a review, and the ground truth review. Your task is to deeply analyze the intents and ground truth very carefully and generate a set of rules that you think are very important to fully capture the nuances of the ground truth review. While analyzing the review please consider the following factors: the exact length of the review, the tone, writing style, structure, important phrases, specific product details, ratings (if any), personal anecdotes, comparisons, and anything else that you think is very important. First start by analyzing the ground truth review extremely carefully accounting for all the important factors within <thinking></thinking> tokens. Once you have done that list a set of extremely detailed and granular rules to ensure that all nuances of the ground truth review are captured to ensure that the generated review is exactly the ground truth review. Include everything including phrases that are important and all stylistic information that needs to be captured in extreme detail. Please enclose these extremely detailed, specific, and granular set of rules within <rules></rules> tags.
\end{lstlisting}
\textbf{With Baseline Rule Generation}
\begin{lstlisting}
You are an expert rule generator whose task is to ensure that a base review can be transformed into the ground truth review. You are provided with the following: The intents that were used to generate the base review, the base review and the ground truth review. You must analyze the differences between the base and the ground truth in great detail analyzing every difference. You must focus on the following while generating these rules: the difference in the length of the reviews, the tone, style, structure, common phrases, specific product details, ratings (if any), personal anecdotes, comparisons, and anything else that you think is very important. All these factors must be closely analyzed to generate these extremely granular set of rules. Please also mention exactly how long the review should be and generate an extremely detailed and granular set of rules that should be able to transform the base review exactly into the ground truth review. To do this please first think deeply and analyze these differences within <thinking></thinking> tags where you can enlist every possible difference between the base and the ground truth review. Once this is done please generate an extremely detailed and granular set of rules that can be used to transform the base review. Do not mention the ground truth or base review in your set of rules whatsoever. The rules should be an extremely detailed guideline to transform the base to ground truth review. Please include these detailed set of rules within <rules></rules> tags.
\end{lstlisting}

\label{prompt:sys-eval}
\subsection{System Prompt: Evaluate Winner}
\subsubsection{Enron Email Dataset}

\begin{lstlisting}
You are an expert email evaluator. Given a number of candidate emails and the ground truth email, your task is to pick which one of the candidate emails is closest to the ground truth email. During your evaluation, please focus mainly on  elements of the email like style, tone, common phrases used, length of the emails, factual accuracy, etc. YOU MUST ALWAYS PICK A WINNER.

Here is how your evaluation should look like:
<evaluation>
<!-- Use this to evaluate each candidate email and compare it with the ground truth -->
</evaluation>
<winner>
<!-- Use this pick the winning candidate email. Display the option that is closest to the ground truth. ONLY DISPLAY THE OPTION NUMBER HERE. For example if email_x is the winner, display only x  -->
</winner>
\end{lstlisting}

\subsubsection{New Yorker Dataset}

\begin{lstlisting}
You are an expert article evaluator. Given a number of candidate articles and the ground truth article, your task is to pick which one of the candidate articles is closest to the ground truth article. During your evaluation, please focus mainly on elements of the article like style, tone, common phrases used, length of the articles, factual accuracy, etc. YOU MUST ALWAYS PICK A WINNER.

Here is how your evaluation should look like:
<evaluation>
<!-- Use this to evaluate each candidate article and compare it with the ground truth -->
</evaluation>
<winner>
<!-- Use this pick the winning candidate article. Display the option that is closest to the ground truth. ONLY DISPLAY THE OPTION NUMBER HERE. For example if article_x is the winner, display only x  -->
</winner>
\end{lstlisting}

\subsubsection{Lamp3U Review Dataset}
\begin{lstlisting}
You are an expert review evaluator. Given a number of candidate reviews and the ground truth review, your task is to pick which one of the candidate reviews is closest to the ground truth review. During your evaluation, please focus mainly on elements of the review like style, tone, common phrases used, length of the reviews, factual accuracy, product details, personal experiences, ratings (if any), comparisons, and overall sentiment. YOU MUST ALWAYS PICK A WINNER.

Here is how your evaluation should look like:
<evaluation>
<!-- Use this to evaluate each candidate review and compare it with the ground truth -->
</evaluation>
<winner>
<!-- Use this pick the winning candidate review. Display the option that is closest to the ground truth. ONLY DISPLAY THE OPTION NUMBER HERE. For example if review_x is the winner, display only x  -->
</winner>
\end{lstlisting}
\section{Human Evaluation}
\label{appendix:human-eval}
\subsection{Human Study Details}
\begin{figure*}[!ht]
    \centering
    \includegraphics[width = \linewidth]{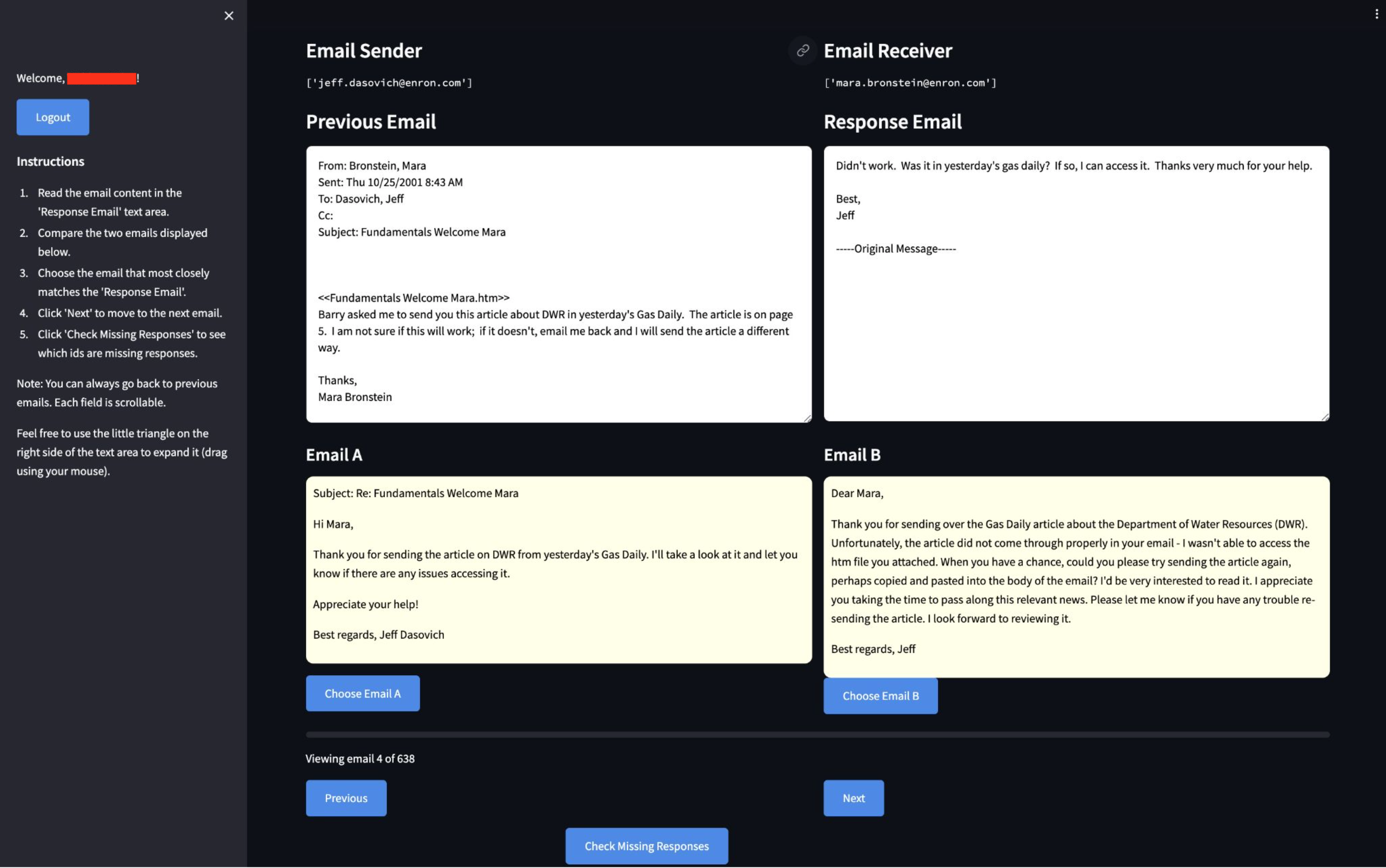}
    \caption{\textbf{Human Evaluator View:} The evaluation screen - including instructions - provided to our human evaluators}
    \label{fig:human-eval-tool}
\end{figure*}

To validate our usage of GPT-4o as an evaluator, we collect human preference data for the same matchups presented to GPT-4o. As seen in Fig \ref{fig:human-eval-tool}, every human evaluator is provided with clear and specific instructions alongside the ground truth. Evaluators are asked to select which of the two options best matches the ground truth. To mitigate biases, all evaluators receive the prompts in the same order and are allowed to review and make changes if needed. We randomly sample 200 comparison examples of our work vs naive finetuning and our work vs no baseline rules alongside 100 comparison examples of our work vs small and large baselines. The same set of human evaluators reviewed and made choices for each subset. We remove missing judgments (which amount to $<$ 1\% of collected data) and measure the raw agreement percentage between humans on the same subset followed by the agreement between each human and GPT-4o.

\subsection{Human Evaluation: Demographics}
We enlisted 50 volunteer human raters, all of whom are pursuing or have obtained degrees in either STEM or business-adjacent fields. The demographic breakdown of our participants is as follows:

\begin{itemize}
    \item \textbf{Gender}: 68\% (34) of our participants are men, while 32\% (16) are women.
    \item \textbf{Age}: The age range of the participants spans from 22 to 50 years, with a median age of 28 years.
    \item \textbf{Education Level}:
    \begin{itemize}
        \item 70\% (35 participants) hold a Bachelor's degree
        \item 20\% (10 participants) have obtained a Master's degree
        \item 10\% (5 participants) have completed or are currently pursuing a Ph.D.
    \end{itemize}
    \item \textbf{Fields of Study}:
    \begin{itemize}
        \item 30\% (15 participants) are from Computer Science or Computer Engineering
        \item 20\% (10 participants) have backgrounds in Engineering (Mechanical, Electrical, Civil, etc.)
        \item 20\% (10 participants) are from Business or Management
        \item 15\% (7 participants) have studied Mathematics or Statistics
        \item 15\% (8 participants) come from various other STEM fields, including Biology, Chemistry, and Physics
    \end{itemize}
\end{itemize}

All volunteers were thoroughly briefed on the goals of this work and provided informed consent for data collection and its subsequent publication. The diversity in their educational and professional backgrounds ensures a comprehensive and balanced evaluation of our research.

We confirm that our study has received IRB approval from our institution for conducting annotations and evaluations of this nature. Our lab has an existing IRB review that covers this type of annotation work for evaluation of this form. All annotators were fully briefed on the study's purpose and provided informed consent for anonymized data release. We do not collect or release any personally identifiable information (PII). For all datasets, we plan to release only anonymized, binary preference information to support future research efforts. This study did not receive federal funding.

We believe that these safeguards ensure the ethical conduct of our research and the responsible handling of data. We are committed to upholding the highest standards of research integrity and protecting the privacy and rights of all individuals involved in our study.

\subsection{Human Evaluation: Instruction Set}

\begin{lstlisting}[language=,breaklines=true,columns=flexible,linewidth=\columnwidth]
1. Read the email content in the "Response Email" text area.
2. Compare the two emails displayed below.
3. Choose the email that most closely matches the "Response Email".
4. Click "Next" to move to the next email.
5. Click "Check Missing Responses" to see which ids are missing responses.
\end{lstlisting}

\section{Qualitative Examples with GPT-4o \& Human Judgements}
\label{appendix:qualitative_examples}
In this section, we present some qualitative examples of GPT-4o evaluation on preference agents, and some examples of personalization per user, alongside the aggregated human verdict. 
\clearpage

\begin{table*}[b!]
    \centering
    \scalebox{0.9}{
        \begin{tabularx}{\textwidth}{>{\raggedright\arraybackslash}X >{\raggedright\arraybackslash}X}
            \toprule
            \textbf{Input} & Given the metadata of the email, previous context, and the intent of the user, please generate an email.

            \textbf{Metadata:} From: benjamin.rogers@enron.com To: berney.aucoin@enron.com Date: Wed, 2 Feb 2000
            
            \textbf{User Intent:}
            \begin{itemize}
                \item I can offer assistance or information about St. Cecilia's kindergarten class regarding your daughter through \hlgreen{my wife, Megan, who is a second-grade teacher at the School}.
                \item She can talk to someone or answer any questions you may have.
            \end{itemize} \\

            \midrule
            \textbf{Ground Truth Email} & In reference to your daughter and St. Cecilia's kindergarten class, let me know if \hlgreen{my wife, Megan, who is a second grade teacher at the School}, can talk to someone or answer any questions for you.. Ben  \\ 
            \midrule
            \textbf{Naive Finetune} &  Bernie: Megan is a 2nd Grade teacher at St. Cecilia's and \hlred{she teaches the Kindergarten class at that school}. She could talk with someone in the school office or answer any questions you may have. Megan can give you the information you need. Please call her to see if she can help with any of these. Megan is Mrs. M. Rogers @ 776-4853. If you happen to talk to her - let her know that Ben Rogers is her husband. Talk to you later. Ben\\ 
            \midrule
            \textbf{Preference Agent} &  Hey Berney, Megan, \hlgreen{my wife who teaches second grade at St. Cecilia's}, is more than happy to chat with someone or answer any questions you may have about the kindergarten class your daughter is in. Let me know if that's something you'd be interested in. Ben\\ 
            \midrule
            \textbf{GPT-4o Judgement} &  
            
            \textbf{Email 1:} The tone here is informal but introduces a formal element by providing a phone number and specific instructions on contacting Megan, which adds complexity not present in the ground truth and \hlred{misclassifies Megan as a kindergarten teacher}. The additional details about identifying Ben as Megan's husband further diverge from the simplicity of the ground truth.
            
            \textbf{Email 2:} This email maintains a similar informal and personal tone. \hlgreen{It correctly and concisely mentions Megan's role} and how she can assist, aligning well with the ground truth.
            
            \textbf{Winner:} 2
            \\ 
            \bottomrule
        \end{tabularx}
    }
    \caption{Preference Agents vs Naive Finetune \textbf{(Hallucinating Factual Information)}}
    \label{tab:pref-agents-hallucination}
\end{table*}

\begin{table*}[ht]
    \centering
    \scalebox{0.9}{
        \begin{tabularx}{\textwidth}{>{\raggedright\arraybackslash}X >{\raggedright\arraybackslash}X}
            \toprule
            \textbf{Input} & Given the metadata of the email, the previous context, and the intent of the user, please generate an email.

            \textbf{Metadata:} From: benjamin.rogers@enron.com To: bryan.kimzey@csfb.com Date: Thu, 9 Nov 2000 
            
            \textbf{Previous Context:} From: bryan.kimzey@csfb.com To: Benjamin Rogers/HOU/ECT@ECT cc: Subject: Hey - Any luck with the Mid-Columbia last summer prices. Thanks Ben
            
            \textbf{Intent:}
            \begin{itemize}
                \item providing the Mid-Columbia last summer prices as requested.
            \end{itemize} \\

            \midrule
            \textbf{Ground Truth Email} &  
            Here you go! Let me know if you need anything else. Ben \\ 
            \midrule
            \textbf{Preference Agent} & 
            Here are the Mid-Columbia last summer prices! Let me know if you need anything else. Ben\\ 
            \midrule
            \textbf{Large Model baseline} & 
            \hlred{Dear Bryan, I wanted to follow up on your request for the Mid-Columbia last summer prices}. I've managed to obtain the information you were looking for. Below are the prices you requested. Let me know if you need anything else. \hlred{Please don't hesitate to reach out} if you have further questions or require additional information. Best regards, Benjamin \\ 
            \midrule
            \textbf{GPT-4o Judgement} &  
            \textbf{Email 1:} Simple and to the point, directly stating what was provided and offering further assistance, similar to the ground truth’s direct offer of help without additional fluff. Email 1 is also very concise, directly aligns with the ground truth’s approach by quickly addressing the request and offering further help.
            
            \textbf{Email 2:} Contains formal phrases such as \hlred{"Dear Bryan", "I wanted to follow up" and "Please don't hesitate to reach out,"} which add a level of formality not present in the ground truth. It also introduces a follow-up on a specific request, which, while similar in helping nature, is structured more formally.

            \textbf{Winner:} 1
            \\ 
            \bottomrule
        \end{tabularx}
    }
    \caption{Preference Agents vs Large Model Baseline \textbf{(Generic Formatting)}}
    \label{tab:pref-agents-generic}
\end{table*}

\begin{table*}[ht]
    \centering
    \scalebox{0.9}{
        \begin{tabularx}{\textwidth}{>{\raggedright\arraybackslash}X >{\raggedright\arraybackslash}X}
            \toprule
            \textbf{Input} & Given the metadata of the email, the previous context, and the intent of the user, please generate an email.

            \textbf{Metadata:} From: benjamin.rogers@enron.com To: mike.miller@enron.com Date: Wed, 14 Jun 2000

            \textbf{Intent:}
            \begin{itemize}
                \item I have worked on the following deals in the first half of 2000.
            \end{itemize} \\
            \midrule
            \textbf{Ground Truth Email} &  
                Mike: \hlgreen{Here is a list of deals I have worked on for the first half of 2000}. Please let me know if you need any other information. Thanks! Ben
            \\ 
            \midrule
            \textbf{Preference Agent} & 
            Hi Mike! \hlgreen{Here's a list of deals I've worked on for the first half of 2000} - let me know if you need anything else! Regards, Ben
            \\ 
            \midrule
            \textbf{Small Model baseline} & 
            Hey Mike, As you know, the first half of 2000 has been a busy time for me, with several deals in the works. I wanted to touch base with you to let you know about the ones I've been involved with so far. \hlred{Those include... [Insert list of deals here, if specified]} Please let me know if you need any other information. Best, Benjamin
            \\ 
            \midrule
            \textbf{GPT-4o Judgement} &  
            \textbf{Email 1:} Closely matches the ground truth in content, directly presenting the list of deals and offering further assistance without additional context or preamble.
            
            \textbf{Email 2:} This email has a more conversational and slightly formal tone, starting with a friendly greeting, adding unnecessary details and an elaborate introduction before mentioning the deals. \hlred{It adds the phrase "Those include... [Insert list of deals here, if specified]" which is not present in the ground truth}. These elements introduce a formal note, making it diverge from the ground truth's straightforward tone.

            \textbf{Winner:} 1
            \\ 
            \bottomrule
        \end{tabularx}
    }
    \caption{Preference Agents vs Small Model Baseline \textbf{(Unnecessary Details)}}
    \label{tab:pref-agents-no-reasoning}
\end{table*}

\begin{table*}[ht]
    \centering
    \scalebox{0.9}{
        \begin{tabularx}{\textwidth}{>{\raggedright\arraybackslash}X >{\raggedright\arraybackslash}X}
            \toprule
            \textbf{Input} & Given the metadata of the email, the previous context, and the intent of the user, please generate an email.

            \textbf{Metadata:} From: debra.perlingiere@enron.com To: s..pollan@enron.com Date: Wed, 22 Aug 2001 Subject: RE:
            
            \textbf{Previous Context:} From: Pollan, Sylvia S. Sent: Wednesday, August 22, 2001 4:31 PM To: Perlingiere, Debra Subject: What is the status on getting NG Energy's Contract ready? Sylvia
            
            \textbf{Intent:}
            \begin{itemize}
                \item The contract is finished, I just need an attorney to review it.
                \item It should be ready in the morning.
            \end{itemize}
            \\
            \midrule
            \textbf{Ground Truth (Debra's Email)} &  
            It is finished, I just need an attorney to review it. \hlgreen{It should be ready in the morning}.
            \\ 
            \midrule
            \textbf{Preference Agent (Bill Variant)} &
            \hlred{Hey Sylvia, Thanks for asking. NG Energy's Contract} is finished. I just need an attorney to review it. It should be ready in the morning. Thanks, Debra Perlingiere
            \\ 
            \midrule
            \textbf{Preference Agent (Gerald Variant)} &
            \hlblue{What's the holdup on attorney review??} Can you please let me know \hlblue{which attorney is responsible for reviewing} this contract? Debra
            \\
            \midrule
            \textbf{Preference Agent (Debra Variant)} & 
            Contract's finished, just waiting on an attorney to review it - \hlgreen{should be ready first thing in the morning}.
            \\
            \midrule
            \textbf{Human Annotator Judgement} &  
            \textbf{Bill Variant:} \hlred{Overtly polite}, mentions the contract even though \hlred{it isn't present in user intents or the ground truth}, ends email with signoff
            
            \textbf{Gerald Variant:} \hlblue{Takes an angry and urgent tone} which isn't present in ground truth email. \hlblue{Doesn't inform the recipient} when the contract will be ready.

            \textbf{Debra Variant:} \hlgreen{Maintains a consistent and direct tone while getting the main intent across}, Aligns closest with ground truth.
            \\ 
            \bottomrule
        \end{tabularx}
    }
    \caption{Preference Agent For Debra vs Preference Agents For Others}
    \label{tab:personalization-evidence1}
\end{table*}

\begin{table*}[ht]
    \centering
    \scalebox{0.9}{
        \begin{tabularx}{\textwidth}{>{\raggedright\arraybackslash}X >{\raggedright\arraybackslash}X}
            \toprule
            \textbf{Input} & Given the metadata of the email, the previous context, and the intent of the user, please generate an email.

            \textbf{Metadata:} From: gerald.nemec@enron.com To: ned.higgins@enron.com Date: Thu, 6 Sep 2001 14:07:59 -0700 (PDT) Subject: Illinois Power CA
            
            \textbf{Previous Context:} 
            
            \textbf{Intent:}
            \begin{itemize}
                \item I am fine with Illinois Power CA's changes.
                \item I have attached a clean version of the document reflecting their changes.
                \item Ask Illinois Power CA to print out two copies and execute them.
            \end{itemize} \\

            \midrule
            \textbf{Ground Truth (Gerald's Email)} &  
            Ned, I am fine with all of their changes. Attached is a clean version reflects their changes. Ask them to print out two copies and execute.
            \\ 
            \midrule
            \textbf{Preference Agent (Bill Variant)} &
            \hlred{Hi Ned, We agree with the changes.} I have attached a clean version that reflects their changes. \hlred{We would appreciate} you sending them an email requesting that they print out two copies and execute them. Thanks, Gerald Nemec
            \\ 
            \midrule
            \textbf{Preference Agent (Gerald Variant)} &
            Fine with the changes. attached a clean version of the Illinois Power CA that reflects their changes. Ask them to print out two copies and execute them.
            \\
            \midrule
            \textbf{Preference Agent (Benjamin Variant)} & 
            \hl{I am fine with all of their changes. , Gerald, 713-345-4169}
            \\
            \midrule
            \textbf{Human Annotator Judgement} &  
            \textbf{Bill Variant:} \hlred{This email utilizes "we" while the ground truth and intent specify "I". The tone is also overtly polite}. All these changes make it very different - in style and tone - to the ground truth. 
            
            \textbf{Gerald Variant:} Short, concise and \hlgreen{doesn't add unnecessary fluff}, aligning with the ground truth.

            \textbf{Benjamin Variant:} While this is concise, \hl{it misses out on 2/3 rds of the intents} and \hl{adds an uneccesary number}.
            \\ 
            \bottomrule
        \end{tabularx}
    }
    \caption{Preference Agent For Bill vs Preference Agents For Others}
    \label{tab:personalization-evidence2}
\end{table*}

\begin{table*}[ht]
    \centering
    \scalebox{0.9}{
        \begin{tabularx}{\textwidth}{>{\raggedright\arraybackslash}X >{\raggedright\arraybackslash}X}
            \toprule
            \textbf{Input} & Given the metadata of the email, the previous context, and the intent of the user, please generate an email.

            \textbf{Metadata:} From: bill.williams@enron.com To: kathryn.sheppard@enron.com Date: Fri, 19 Oct 2001 Subject: RE: Meeting with Britt Whitman
            
            \textbf{Previous Context:} From: Sheppard, Kathryn Sent: Friday, October 19, 2001 10:02 AM To: Williams III, Bill Subject: Meeting with Britt Whitman Bill, Would you be available to meet with Britt Whitman, Houston Competitive Analysis, at 11:00 a.m. on Tuesday, 10/23? Thanks, Kathy
            
            \textbf{Intent:}
            \begin{itemize}
                \item I will be available on Tuesday.
                \item I could meet with Britt at 3 PM or at 9 AM.
                \item What is the purpose of this meeting?
            \end{itemize} \\

            \midrule
            \textbf{Ground Truth (Bill's Email)} &  
            Kathy, I will be available on Tuesday but it would have to be either earlier or later on in the day. I could meet with Britt at 3PM or at 9 AM. Any idea what this is regarding? Thanks, Bill
            \\ 
            \midrule
            \textbf{Preference Agent (Bill Variant)} &
            Hi Kathy, should be available earlier or later in the day on Tuesday - \hlgreen{how about 9am or 3pm?} What's this meeting about? Thanks, Bill
            \\ 
            \midrule
            \textbf{Preference Agent (Gerald Variant)} &
            Kathy, I will be available on Tuesday, but it would have to be either earlier in the day or later in the day. I could meet with you at 1. 9 AM or 2. 3 PM.
            \\
            \midrule
            \textbf{Preference Agent (Dutch Variant)} & 
            hey kathy, tuesday works for me, but not at 11am. \hlred{my schedule is pretty open}, so how about 9am or 3pm instead? \hlred{not that i need to know too much about what we're discussing}, but do you have any idea what britt wants to chat about? let me know when works for you, bill
            \\
            \midrule
            \textbf{Human Annotator Judgement} &  
            \textbf{Bill Variant:} \hlgreen{concise - even reorders the proposed reschedule timings} to be AM then PM. 
            
            \textbf{Gerald Variant:} concise and clear \hlblue{but doesn't include the intent which asks for the purpose of this call}.

            \textbf{Dutch Variant:} \hlred{The tone is friendlier, chattier, and longer than the ground truth}. Diverges because of additional details.
            \\ 
            \bottomrule
        \end{tabularx}
    }
    \caption{Preference Agent For Bill vs Preference Agents For Others}
    \label{tab:personalization-evidence3}
\end{table*}

\begin{table*}[ht]
    \centering
    \scalebox{0.80}{
        \begin{tabularx}{\textwidth}{>{\raggedright\arraybackslash}X >{\raggedright\arraybackslash}X}
            \toprule
            \textbf{Input} & Given the metadata of the email, the previous context, and the intent of the user, please generate an email.

            \textbf{Metadata:} Date: 2016-03-28 Author: None Title: The Fantasies of China’s Generation X
            
            \textbf{Intent:}
            \begin{itemize}
                \item Cao Fei's art exhibition at MOMA PS1 explores the fantasies and entertainments of China's Generation X.
                \item The exhibition includes Cao Fei's 2000 work "Plant Contest", a traditional Chinese performance.
                \item Cao Fei has also videotaped cosplayers in her native Guangzhou as part of the exhibition.
                \item In addition, she has built a virtual city in the online world Second Life, where her avatar, China Tracy, serves as a philosopher-guide.
                \item Cao Fei was born in 1978 and is based in Beijing.
                \item The exhibition opens on April 3 at MOMA PS1.
            \end{itemize} \\

            \midrule
            \textbf{Ground Truth Article} &  
            You may know the ancient Chinese game Go, but the traditional \hlgreen{“Plant Contest”} (above), which Cao Fei staged in 2000, is probably less familiar. The picture is one of the earliest works in the Beijing-based artist’s show at MOMA PS1, opening on \hlgreen{April 3}. Cao, who was born in 1978, delves into the fantasies and entertainments of her generation, videotaping cosplayers in her native Guangzhou and building a virtual city in the online world Second Life, where her avatar, \hlgreen{China Tracy}, has served as a philosopher-guide.
            \\ 
            \midrule
            \textbf{Preference Agent} & 
            The Fantasies of China's Generation X Cao Fei is set to unveil an immersive and thought-provoking exhibition at MOMA PS1, which delves into the fantasies and entertainments of China's Generation X. \hlgreen{The exhibition, opening on April 3}, promises to be a fascinating exploration of the cultural landscape of a generation that grew up in a changing China. \hlgreen{One of the highlights of the exhibition is Cao Fei's 2000 work, "Plant Contest"}, a traditional Chinese performance that offers a glimpse into the country's rich cultural heritage. This work is complemented by Cao Fei's videotapes of cosplayers in her native Guangzhou, which provide a window into the lives of young Chinese people and their fascination with fantasy and role-playing. Cao Fei has also created a virtual city in the online world Second Life, where her avatar, \hlgreen{China Tracy}, serves as a philosopher-guide. 
            \\ 
            \bottomrule
        \end{tabularx}
    }
    \caption{Preference Agents vs Naive Finetune (Page 1 of 2)}
    \label{tab:pref-agents-newyorker}
\end{table*}

\begin{table*}[ht]
    \centering
    \scalebox{0.9}{
        \begin{tabularx}{\textwidth}{>{\raggedright\arraybackslash}X >{\raggedright\arraybackslash}X}
            \toprule
            \textbf{Naive Finetune} & 
            \hlred{In China, it’s often said that each generation is a socialist addendum. The Millennials are called the HelloKittyState Generation}. Cao Fei was born in 1978, smack in the middle of China’s Generation X. Her \hlred{“Triple Artist” show} at MOMA PS1 (\hlred{April 3-June 5}) includes a work she made in 2000, a sort of Fluxus performance in a Guangzhou barber shop, in which participants award her eggplants for proper beards. Last year she was videotaping her own local cosplayers. Cao Fei (based in Beijing) has also spent time building her own virtual city in the online world of Second Life: China Tracy serves as a philosopher-guide.
            \\ 
            \midrule
            \textbf{GPT-4o Judgement} &  
             \textbf{Article 1} is a clear winner due to its similarities in style, tone, and factual accuracy with the ground truth article. \hlgreen{The structure and organization of the articles are also similar}, starting with an introduction to the artist and the exhibition, followed by a description of the various works featured. <...>.
             
             \textbf{Article 2} has a more informal tone and uses colloquial language, which differs from the ground truth article. The structure of the article is also \hlred{less cohesive, jumping between different topics and ideas without a clear flow}. It also mentions the exhibit \hlred{ending on June 5th which isn't present in the ground truth}. While it does mention some of the same works as the ground truth article, the descriptions are brief and lack the detail and context provided in Candidate Article 1.

             \textbf{Winner:} 1
            \\ 
            \midrule
            \textbf{Human Choice:} & 1
            \\
            \bottomrule
        \end{tabularx}
    }
    \caption{Preference Agents vs Naive Finetune (Page 2 of 2)}
    \label{tab:pref-agents-newyorker-2}
\end{table*}

\end{document}